\documentclass[10pt,twocolumn,letterpaper]{article}

\usepackage{iccv}
\usepackage{times}
\usepackage{epsfig}
\usepackage{graphicx}
\usepackage{amsmath}
\usepackage{amssymb}
\usepackage{subcaption}
\usepackage{stfloats}

\usepackage[breaklinks=true,bookmarks=false]{hyperref}

\iccvfinalcopy %

\def\iccvPaperID{6} %

\usepackage{multirow}
\usepackage{booktabs}
\usepackage{pifont}
\usepackage{diagbox}
\newcommand{\cmark}{\ding{51}}%
\newcommand{\xmark}{\ding{55}}%

\usepackage[table]{xcolor}
\definecolor{pinkcarino}{RGB}{255,0,255}

\definecolor{road}{RGB}{128,  64, 128}
\definecolor{nature}{RGB}{107, 142,  35}
\definecolor{person}{RGB}{220,  20,  60}
\definecolor{vehicle}{RGB}{  0,   0, 142}
\definecolor{construction}{RGB}{ 70,  70,  70}
\definecolor{obstacle}{RGB}{153, 153, 153}
\definecolor{water}{RGB}{45,  60, 150}
\definecolor{unknown}{RGB}{0,0,0}

\usepackage{tcolorbox}

\usepackage{tabularx}

\newcolumntype{Y}{>{\centering\arraybackslash}X}

\newcommand{\dname}[1][\ ]{\texttt{SynDrone}#1}

\begin{document}

\title{SynDrone -- Multi-modal UAV Dataset for Urban Scenarios}

\author{Giulia Rizzoli* \hspace{1em} Francesco Barbato* \hspace{1em} Matteo Caligiuri* \hspace{1em} Pietro Zanuttigh\\
University of Padova\\
Department of Information Engineering\\
Via Gradenigo 6/b\\
{\tt\small{\{giulia.rizzoli,francesco.barbato,matteo.caligiuri,pietro.zanuttigh\}@dei.unipd.it}}
}

\maketitle
\ificcvfinal\thispagestyle{empty}\fi

\begin{abstract}
   The development of computer vision algorithms for Unmanned Aerial Vehicles (UAVs) imagery heavily relies on the availability of annotated high-resolution aerial data. However, the scarcity of large-scale real datasets with pixel-level annotations poses a significant challenge to researchers as the limited number of images in existing datasets hinders the effectiveness of deep learning models that require a large amount of training data.  In this paper, we propose a multimodal synthetic dataset containing both images and 3D data taken at multiple flying heights to address these limitations. In addition to object-level annotations, the provided data also include pixel-level labeling in 28 classes, enabling exploration of the potential advantages in tasks like semantic segmentation. In total, our dataset contains 72k labeled samples that allow for effective training of deep architectures showing promising results in synthetic-to-real adaptation. The dataset will be made publicly available to support the development of novel computer vision methods targeting UAV applications.
\end{abstract}

\begin{figure}[htp]
    \setlength{\fboxsep}{2pt}
    \centering    
    \includegraphics[width=0.475\textwidth]{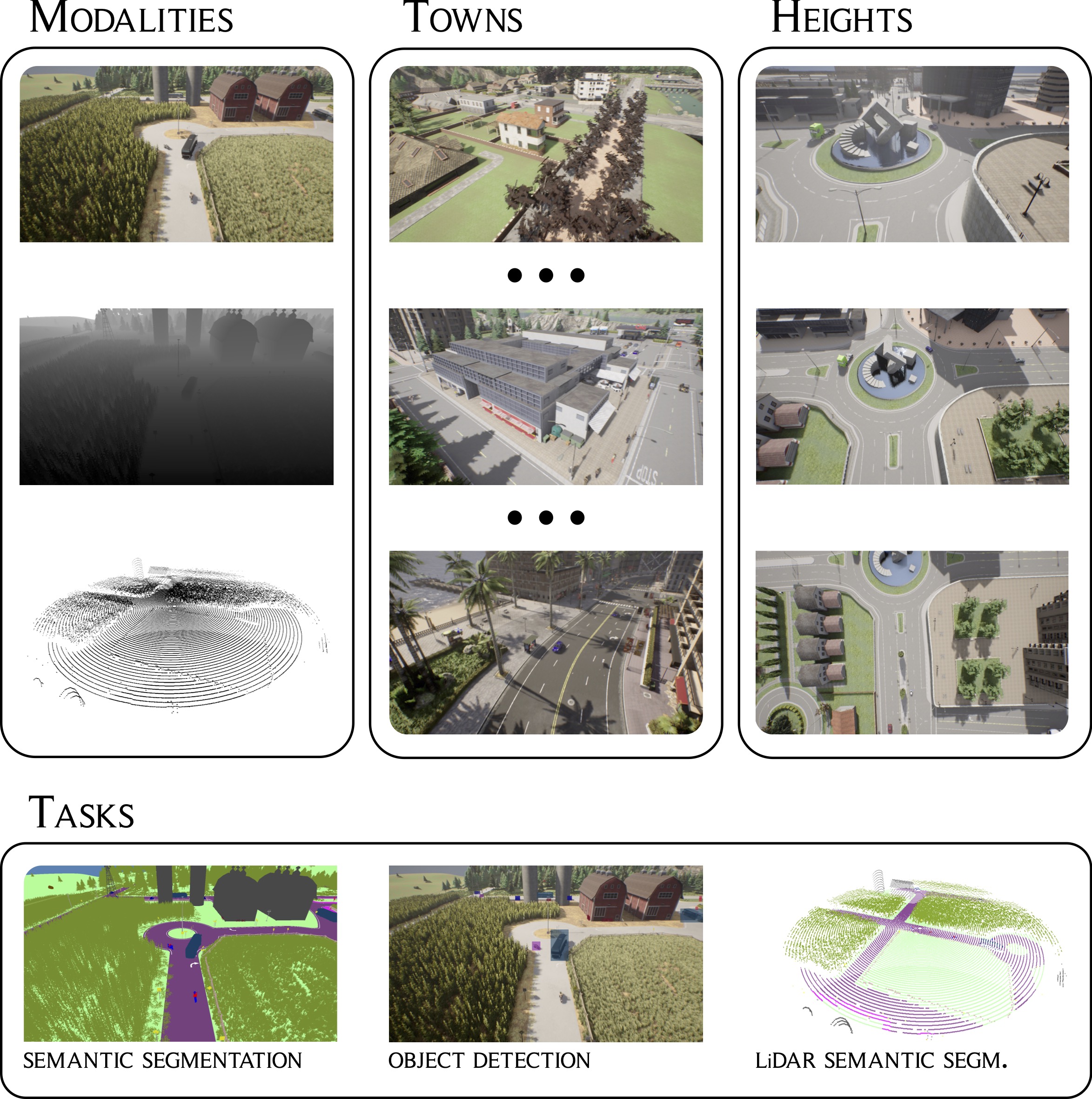}
    \caption{\dname is a comprehensive dataset aggregating multi-sensor, multi-location, and multi-altitude data with different task-specific labeling. %
    }
    \label{fig:abstract}
\end{figure}

\section{Introduction}
Unmanned aerial vehicles (UAVs) have revolutionized various applications, including surveillance \cite{motlagh2017uav, kim2018designing}, monitoring \cite{elloumi2018monitoring, de2019multi}, agriculture \cite{bhatnagar2022mapping}, and mapping \cite{chen2021reconstruction}. In particular, UAVs have shown great potential in urban scene analysis, enabling tasks such as traffic control \cite{elloumi2018monitoring}, population assessment \cite{robicquet2016learning, motlagh2017uav}, urban greenery maintenance \cite{behera2022vegetation}, and road marking extraction \cite{guan2022road}. Despite the availability of UAV datasets for detection, tracking, and behavior analysis, there remains a lack of comprehensive datasets specifically designed for densely-annotated tasks such as semantic segmentation. 
The existing UAV semantic segmentation datasets \cite{nigam2018ensemble,lyu2020uavid,chen2018large,9015998} have limitations in terms of size, scene variation, sampling rate, and class labeling set. Moreover, most of them do not account for the multiple possible operating altitudes and camera angles. 
To address these limitations, we present a new synthetic drone imagery dataset, called \textbf{\dname[]}\footnote{The code used for analyses and generation of the dataset, and links for its download are available at \href{https://github.com/LTTM/Syndrone}{{https://github.com/LTTM/Syndrone}}.}%
, which contains significantly larger image sets that capture scenes with a broader variability and increased complexity. %
In addition to traditional remote sensing images, UAVs equipped with Light Detection and Ranging (LiDAR) and depth cameras have emerged as powerful tools for high-precision positioning systems \cite{guan2022road}. These UAVs can dynamically explore uncharted territories, capturing both 3D scans and aerial images in real-time. This capability enables the development of advanced vision tasks, such as height map generation and 3D scene understanding, which can contribute to the accurate analysis and interpretation of the surrounding environment.
To this extent, together with standard RGB images \dname includes co-registered depth maps and LiDAR data, providing the capability for multi-modal analysis as well as supporting other tasks such as depth estimation, navigation, and 3D reconstruction.
Moreover, \dname also enables other vision tasks such as object detection and instance tracking. The dataset includes bounding box annotations and object identifiers, making it suitable for multi-tasking approaches. %

\noindent In summary, we introduce several  contributions:
\begin{enumerate}
    \item We propose \textbf{\dname[]}%
    , a synthetic large-scale multi-modal dataset providing drone imagery with a high-sampling rate at variable altitudes and view angles. 
    \item We provide registered color, depth, and LiDAR data allowing the development of multi-modal schemes.
    \item We provide ground truth data for multiple tasks, including bounding boxes, object class labels, and semantic segmentation labeling, hence the dataset is suitable for object detection and tracking but also for more advanced pixel-level recognition applications.
    \item We showcase benchmark results with different models and evaluate the suitability of the data for transfer learning by testing the trained models on real-world datasets.
\end{enumerate}

\section{Related Work}
In this section, we provide a comprehensive overview of existing methods and datasets focusing on drone imagery in image-level tasks within urban scenarios. A summary of the datasets for object-level tasks can be found in Table \ref{tab:obj-level}, while in Table \ref{tab:px-level} is reported the equivalent for pixel-level tasks. 
Notice that we focus on works targeting drone-level applications (\ie, with a flying altitude below 100m), there is also a wide body of research tackling satellite or high-altitude flying data that represents a different field.

\subsection{Vision in UAV urban scenarios}

Drone technology has witnessed significant advancements in recent years, revolutionizing urban planning and management. The intersection of urban scene analysis tasks and computer vision encompasses various subdomains, including
detection \cite{balamuralidhar2021multeye, cai2018cascade, lin2017focal}, trajectory prediction \cite{he2022irlsot, zhou2019omni, wang2018learning, luo2019bag}, depth estimation \cite{madhuanand2021self, pirvu2021depth, miclea2021monocular}, and semantic segmentation \cite{he2022swin, wang2021ranpaste, li20222dsegformer, marcu2020semantics}.
Furthermore, the availability of data has garnered significant interest from the %
community, leading to the exploration of various learning frameworks such as Continual Learning \cite{marsocci2023continual}, Cross-Domain Learning \cite{li2021learning,zhang2021curriculum}, Few-shot Learning \cite{khoshboresh20222d,patsiouras2020few} and Multi-modal Learning \cite{speth2022deep}.
Ultimately, in response to the limitations and challenges associated with UAVs and their operations, researchers have directed their efforts towards addressing critical issues, \eg, adversarial attacks \cite{wang2023defense, wang2023global}, and to the development of lightweight  architectures \cite{liu2021light}. 

\begin{table*}[!t]
\centering
\resizebox{2\columnwidth}{!}{
    \begin{tabular}{cccccccccccc}
    \hline
    Name & Year & Task & S/R & MM & \# classes & \# images & \# sequences & Frequency {[}Hz{]} & Height {[}m{]} & Size {[}px{]} & View Angle \\
    \hline
    Campus \cite{robicquet2016learning} & 2016 & MOT & R  & \xmark & - & 930K & 100+ & - & 80 & 1400x1904 & 90 \\
    DBT70 \cite{li2017visual} & 2017 & SOT & R & \xmark & - & - & 70 & - & - & 1280×720& variable\\
    VisDrone-Img \cite{zhu2021detection} & 2018 & DET & R &  \xmark & 10 & 10209 & - & - & - & 2000x1500 & variable \\
    VisDrone-Vid \cite{zhu2021detection} & 2018 & DET & R & \xmark & 10 & 40k & 96 & - & - & 3840x2160 & variable \\
    VisDrone-SOT \cite{zhu2021detection} & 2018 & SOT & R & \xmark & - & 139.3k & 167 & - & - & - & variable  \\
    VisDrone-MOT \cite{zhu2021detection} & 2018 & MOT & R & \xmark & - & 108.3k & 96 & - & - & 3840x2160 & variable \\
    UAVDT \cite{du2018unmanned} & 2018 & DET, SOT, MOT & R & \xmark & 3 & $\sim$ 80k (37.2k + 40.7k) & 100 & 30 & 10-70+ & 1080×540 & front/side/bird \\
    AU-AIR \cite{bozcan2020air} & 2020 & DET & R & \xmark & 8 & 32823 & 8 & 5 & 5-30 & 1920x1080  &  45 to 90  \\
    MDOT \cite{zhu2020multi} & 2020 & SOT & R & \xmark & 9 & 259793 & 155 & - & 20-100 & 1280×720 & - \\
    UAV123 \cite{mueller2016benchmark} & 2020 & SOT & S+R & \xmark & - & 112578 & 123 & 30 to 96 & 5-25 & 1280×720 to 3840x2160 & - \\
    Anti-UAV \cite{jiang2021anti} & 2021 & SOT & R & \cmark & 1 & 318 & 318 & 25 & - & - & -\\
    HIT-UAV \cite{suo2023hit} & 2023 & DET & R & \cmark & 4 & 2898 & - & 7 & 60-130 & 640×512 & 30 to 90 \\
    \bottomrule
\end{tabular}}
\caption{Object-level UAV datasets. S/R = Synthetic/Real, MM = MultiModal. DET = DETection, SOT = Single Object Tracking, MOT = Multiple Object Tracking. - = not applicable or not explicit in the paper.}
\label{tab:obj-level}
\end{table*}\begin{table*}[!t]
\resizebox{2\columnwidth}{!}{
    \begin{tabular}{ccccccccccc}
    \hline
    Name & Year & MM & BB & \# classes & \# images & \# sequences & Frequency {[}Hz{]} & Height {[}m{]} & Size {[}px{]} & View Angle \\
    \hline
    Aeroscapes \cite{nigam2018ensemble} & 2018 & \xmark & \xmark & 11 & 3269 & 141 & - & 5-50 & 1280x720 & variable\\
    ICG Drone \cite{ICGDroneDataset} & 2018 & \cmark & \xmark & 20 & 400 & - & 1 & 5-30 & 6000x4000 & 90 \\
    UDD \cite{chen2018large} & 2018 & \xmark & \xmark & 4 & 301 & - & - & 60-100 & 4096x2160 or 4000x3000  & variable\\
    UAVid \cite{lyu2020uavid} & 2020 &\xmark %
    & \cmark & 8 & 270 %
    & 30 & 0.2 & 50 & 4096x2160 or 3840x2160 & 45 \\
    \dname (Ours) & 2023 & \cmark & \cmark & 28 & (60+12)k & 24 & 25 & 20, 50, 80 & 1080$\times$1920 & 30, 60, 90 \\
    \bottomrule
    \end{tabular}
}
\caption{Pixel-level UAV datasets. BB = Bounding Boxes. - = not explicit in the paper.}
\label{tab:px-level}
\end{table*}

\subsection{Object-level UAV datasets for urban scenarios}
Object-level drone datasets have a fundamental significance in advancing research and development in various computer vision tasks. %
These datasets provide annotated images and videos captured from unmanned aerial vehicles (UAVs), enabling the training and evaluation of algorithms for object detection, tracking, and other related applications. We summarize several notable object-level drone datasets specifically designed for urban scenarios, highlighting their key %
contributions. Refer to Table \ref{tab:obj-level} for details.

\textbf{Campus} \cite{robicquet2016learning} %
is a large-scale %
dataset designed for multi-object tracking, activity understanding, and trajectory forecasting within the Stanford University campus. The images were captured using a top-down camera mounted on a multirotor drone hovering at a high altitude. %

\textbf{DBT70} \cite{li2017visual} %
consists of 70 video sequences captured from various sources, including drones and YouTube. The dataset contains manually annotated bounding boxes for pedestrians and vehicles. %

\textbf{UAV123} \cite{mueller2016benchmark} %
is a benchmark dataset for UAV tracking tasks, consisting of 100+ video sequences. It encompasses data from professional-grade and consumer-grade UAVs as well as simulator-generated data. %

\textbf{VisDrone} \cite{zhu2021detection} %
is a large-scale benchmark dataset %
containing a significant number of images with the corresponding annotations. The dataset has been expanded and updated over time to include more data and improve its coverage. %
It covers various environmental conditions, such as different weather conditions (\eg, sunny, cloudy, rainy), different altitudes, and camera viewpoints.

\textbf{Anti-UAV} \cite{jiang2021anti} %
includes videos of different UAV types flying in various lighting conditions (day and night), light modes (infrared and visible), and diverse backgrounds. It aims to ensure the diversity of data for tracking purposes.

\textbf{UAVDT} \cite{du2018unmanned} %
contains 100 video sequences captured from a UAV platform in urban areas, including scenes such as highways and T-junctions. %
The dataset offers annotations for tracking tasks, including object-bounding boxes.

\textbf{MDOT} \cite{zhu2020multi} %
is designed for multi-drone single-object tracking. It includes video clips captured by two or three drones simultaneously tracking the same target at different daytime. %

\textbf{AU-AIR} \cite{bozcan2020air} %
comprises of images captured by a multirotor drone flying at low altitudes in an urban scenario. It includes %
bounding boxes for instances of people and vehicles. Moreover, while target IDs are unavailable, the dataset provides image-level metadata, including drone speed, %
latitude, and longitude.

\textbf{HIT-UAV} \cite{suo2023hit} %
is a high-altitude infrared thermal dataset for object detection on UAVs. It contains infrared thermal images extracted from hundreds of videos captured in various scenarios such as schools, parking lots, and playgrounds. The dataset enables the evaluation of object detection algorithms specifically designed for thermal imaging.

\subsection{Pixel-level UAV datasets}
Pixel-level UAV datasets with semantic segmentation annotations play a crucial role in developing and evaluating algorithms for various applications, including autonomous navigation, scene understanding, and 3D reconstruction. In this section, we overview  several pixel-level UAV datasets, highlighting their strengths and limitations (see Table \ref{tab:px-level}).

\textbf{Aeroscapes} \cite{nigam2018ensemble} %
The Aeroscapes dataset stands out by its focus on capturing urban scenes using drones, which enables the collection of more diverse and informative data compared to traditional car-mounted cameras. The dataset includes 11 classes and comprises %
141 video sequences, with images having a resolution of 1280x720 pixels.\\ %
\indent \textbf{ICG Drone} \cite{ICGDroneDataset} %
provides a collection of high-resolution imagery captured from a bird's eye view, facilitating a semantic understanding of residential and green urban scenes. It includes more than 20 houses captured at low altitudes. The dataset offers pixel-accurate annotations for 22 classes, covering a wide range of residential area objects and structures (a notable shortage is the lack of road class). Additionally, it provides valuable supplementary data such as fish-eye stereo images, thermal images, ground control points, and 3D ground truth. %

\indent \textbf{UAVid} \cite{lyu2020uavid} %
distinguishes itself by providing video sequences captured by small UAVs in various locations. This dataset offers labeled images at a lower frame rate (0.2 FPS) and unlabeled images at a higher frame rate (20 FPS). With 30 video sequences comprising a total of 300 labeled images, UAVid offers the opportunity to explore self-supervised learning approaches for semantic segmentation and 3D reconstruction. %

\indent \textbf{Urban Drone Dataset (UDD)} \cite{chen2018large} %
specializes in aiding 3D reconstruction tasks using an improved Structure From Motion (SFM) method. With images captured at altitudes between mid and high altitudes, UDD provides a variety of urban scenes from four different cities in China. The dataset offers annotations for 4 semantic classes. %

Most real-world datasets lack an adequate quantity of images or only focus on short sequences (see Table \ref{tab:px-level}), making it challenging to train a network capable of generalizing well to different data. The majority have a low sampling rate because annotating each frame is prohibitively expensive. This limitation hampers the potential for leveraging video semantic segmentation. Additionally, many datasets have a restricted range of classes or do not specifically emphasize driving-related categories. For this reason, despite having a wide range of classes, the ICG Drone dataset, which notably does not include the road class, has restricted applicability to driving or monitoring scenarios.

Ultimately, while other existing methods for generating large-scale synthetic aerial data \cite{9015998} have been proposed, they lack the capability of simulating relevant dynamic elements such as vehicles and pedestrians.
\section{The \dname Dataset}

\begin{figure*}[t]
    \centering
    \includegraphics[width=1\textwidth]{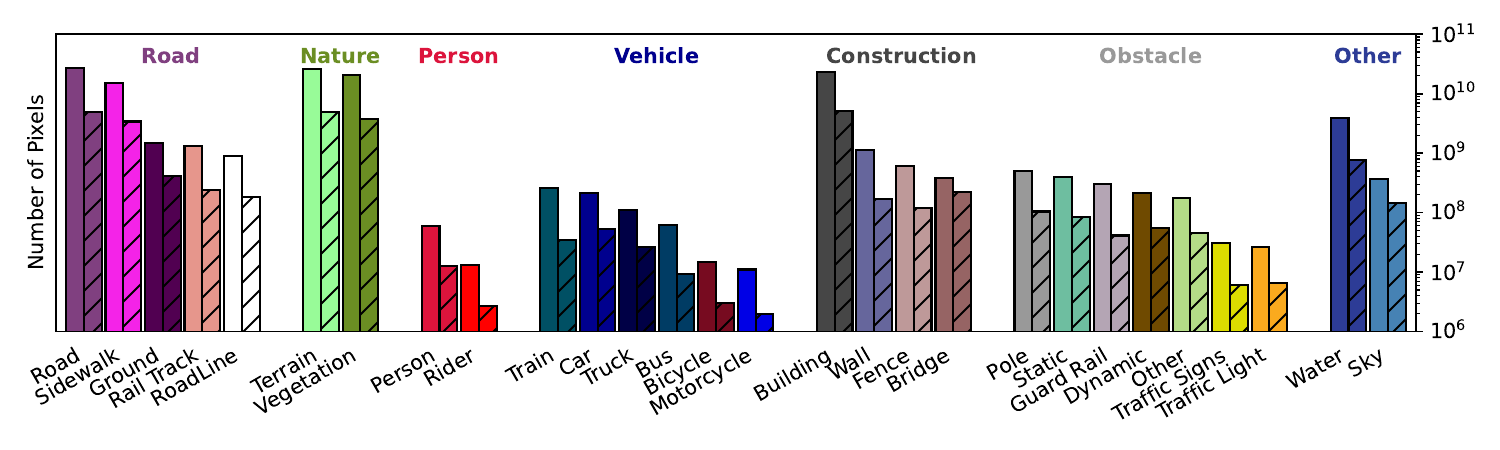}
    \caption{Class distribution in logarithmic scale. Flat bars for the training set, dashed for the test set. Names at the top refer to the coarse grouping, see Table \ref{tab:mapping} for details.}
    \label{fig:classes}
\end{figure*}

In this section, we detail the construction and contents of the proposed \dname dataset. 
It is a multimodal synthetic dataset, developed for the task of drone imagery understanding at both object and pixel-level in urban settings. 
The dataset contains 72k frames captured from drone views, which are grouped into 8 sequences and further split into 60k images for training and 12k images for testing.
The data is densely labeled into 28 semantic classes, with object-level labeling as bounding boxes for the moving objects (vehicles and pedestrians). The data were collected using a modified version of the CARLA simulator \cite{dosovitskiy2017carla,testolina2023selma} (see Section \ref{sec:carla}) at a frequency of 25 Hz and at different heights of 20, 50, and 80 meters above the ground.
The images have a resolution of $1920\times1080$ pixels and are captured from different viewing angles of 30, 60, and 90 degrees w.r.t. the horizontal plane. Further detail on the camera sensors and the acquired trajectories are in Section \ref{sec:setup}. The class distribution is shown in Figure~\ref{fig:classes}.

\subsection{The CARLA simulator}
\label{sec:carla}
We decided to employ the CARLA simulator \cite{dosovitskiy2017carla}, which has been previously used to generate synthetic data in the autonomous driving context \cite{alberti2020idda,testolina2023selma}. 
Built upon Unreal Engine 4 (UE4), CARLA offers high-quality rendering, realistic physics powered by NVIDIA PhysX, and basic Non-Player Character (NPC) logic. %
We employ a modified CARLA 0.9.12 version \cite{testolina2023selma} that 
provides a diverse range of carefully designed UE4 models, encompassing static objects (\eg, buildings, vegetation, traffic signs) and dynamic objects (\eg, vehicles, pedestrians). These models share a common scale and realistic sizes.
The original version includes a blueprint library with 24 car models, 6 truck models, 4 motorbike models, and 3 bike models, each customizable in terms of colors. Additionally, it features 41 pedestrian models of various ethnicities, builds, and attired in a wide array of clothes. Furthermore, CARLA offers 8 meticulously crafted towns (Town01-07 and Town10HD), incorporating over 40 building models. Each town possesses unique features and landmarks, providing 8 simulation environments with distinct visual characteristics.
CARLA facilitates data retrieval from the simulated world through various sensors. These sensors can be precisely positioned, rotated, and attached to parent actors, enabling them to follow rigid or spring-arm-like movements. 
Sensor data can be collected at each simulation step. When using multiple high-resolution sensors, a synchronous mode ensures that the GPU completes rendering and delivers the data to the client before the subsequent simulation step, guaranteeing a consistent sensor acquisition rate across all sensors.
In the modified version, the semantic class set has been extended to ensure compatibility with existing benchmark datasets for autonomous driving \cite{cordts2016cityscapes, geiger2013vision}. 
To allow such extension \cite{testolina2023selma} introduced multiple new vehicle models, such as trains, trams, buses, and trucks. %

\subsection{Acquisition setup}
\begin{figure*}[t]
    \centering
    \begin{subfigure}{.24\textwidth}
        \caption*{Town01}
        \includegraphics[width=\textwidth,height=.7\textwidth]{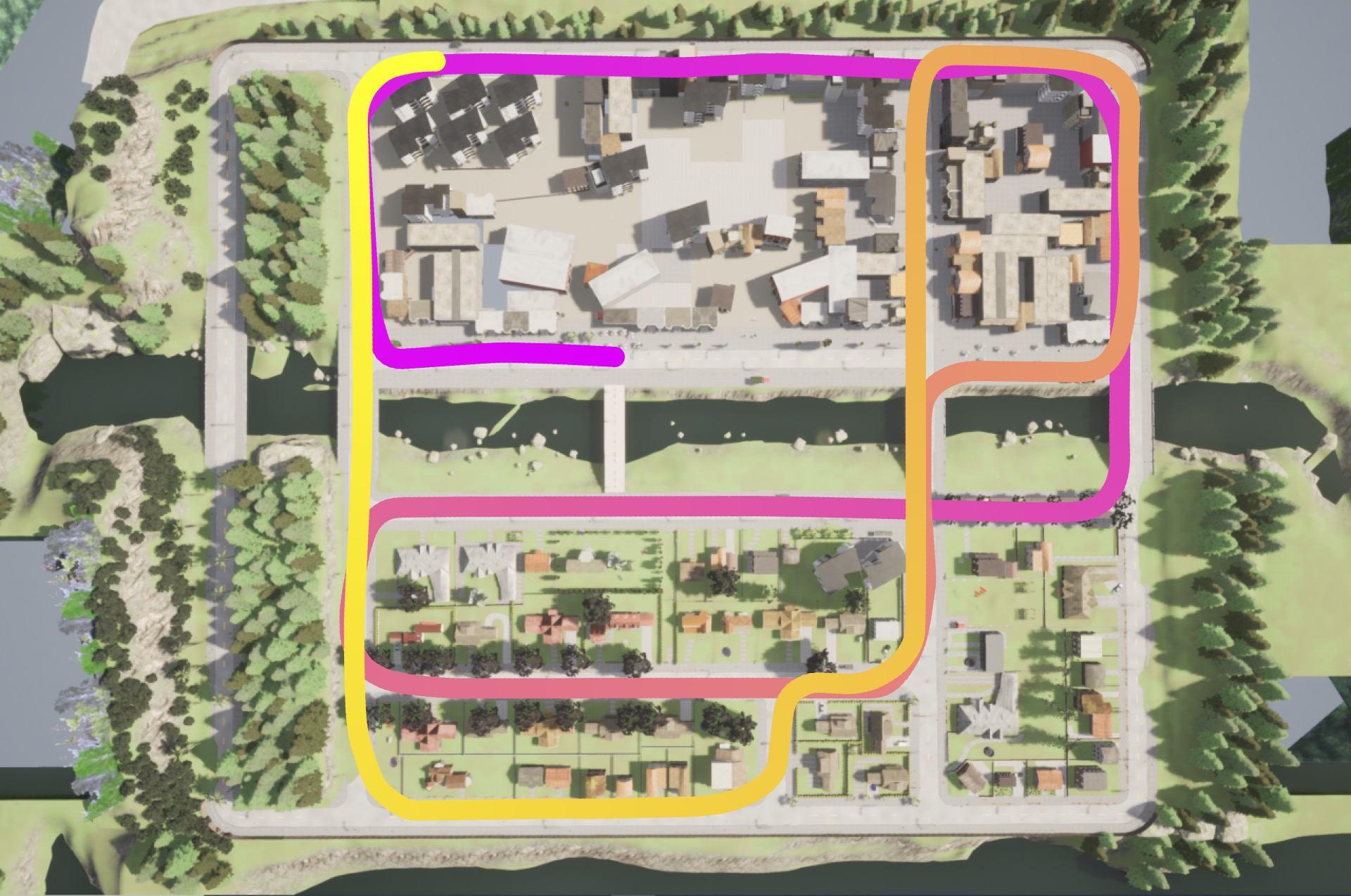}
    \end{subfigure} %
    \begin{subfigure}{.24\textwidth}
        \caption*{Town02}
        \includegraphics[width=\textwidth,height=.7\textwidth]{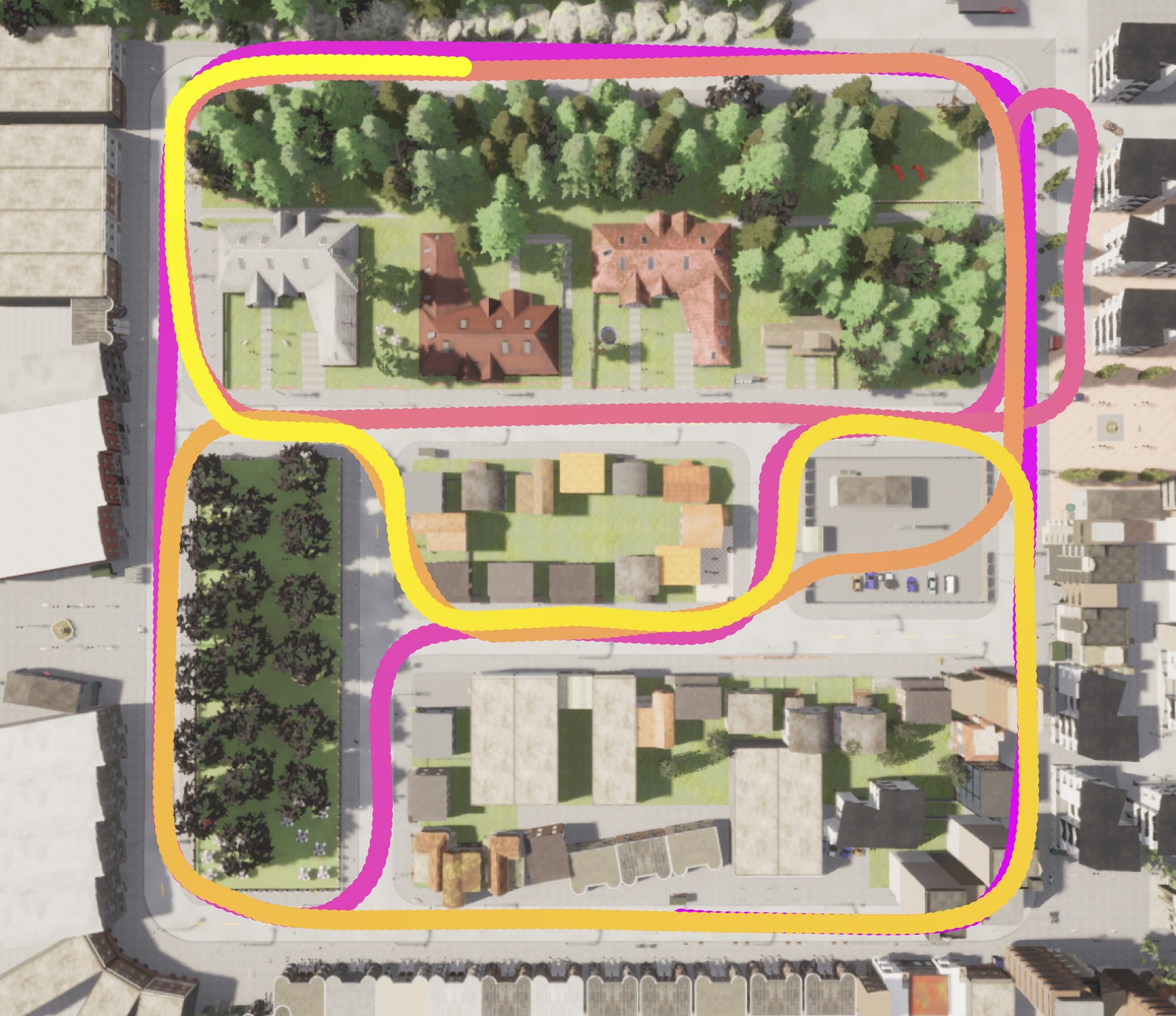}
    \end{subfigure} %
    \begin{subfigure}{.24\textwidth}
        \caption*{Town03}
        \includegraphics[width=\textwidth,height=.7\textwidth]{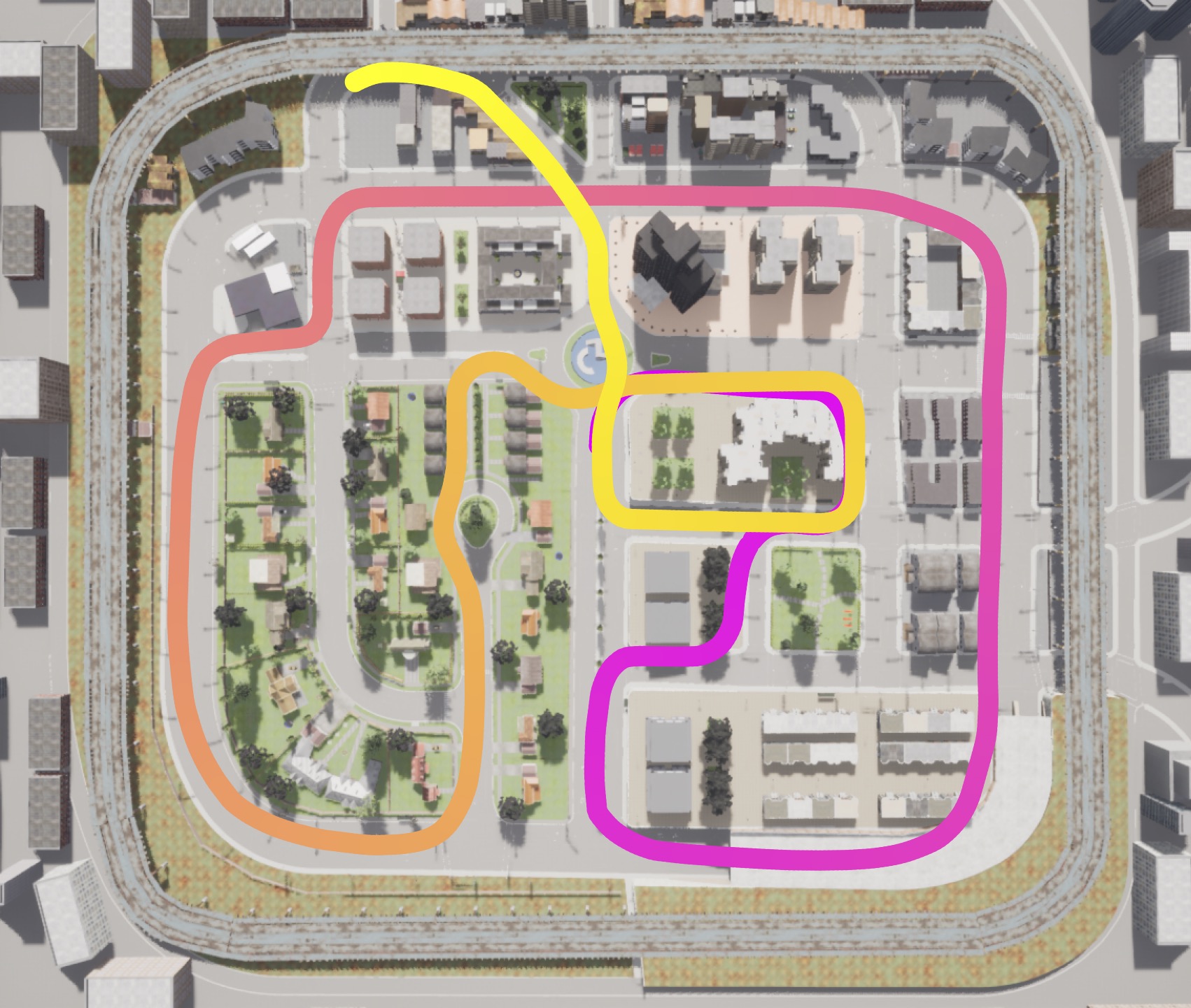}
    \end{subfigure} %
    \begin{subfigure}{.24\textwidth}
        \caption*{Town04}
        \includegraphics[width=\textwidth,height=.7\textwidth]{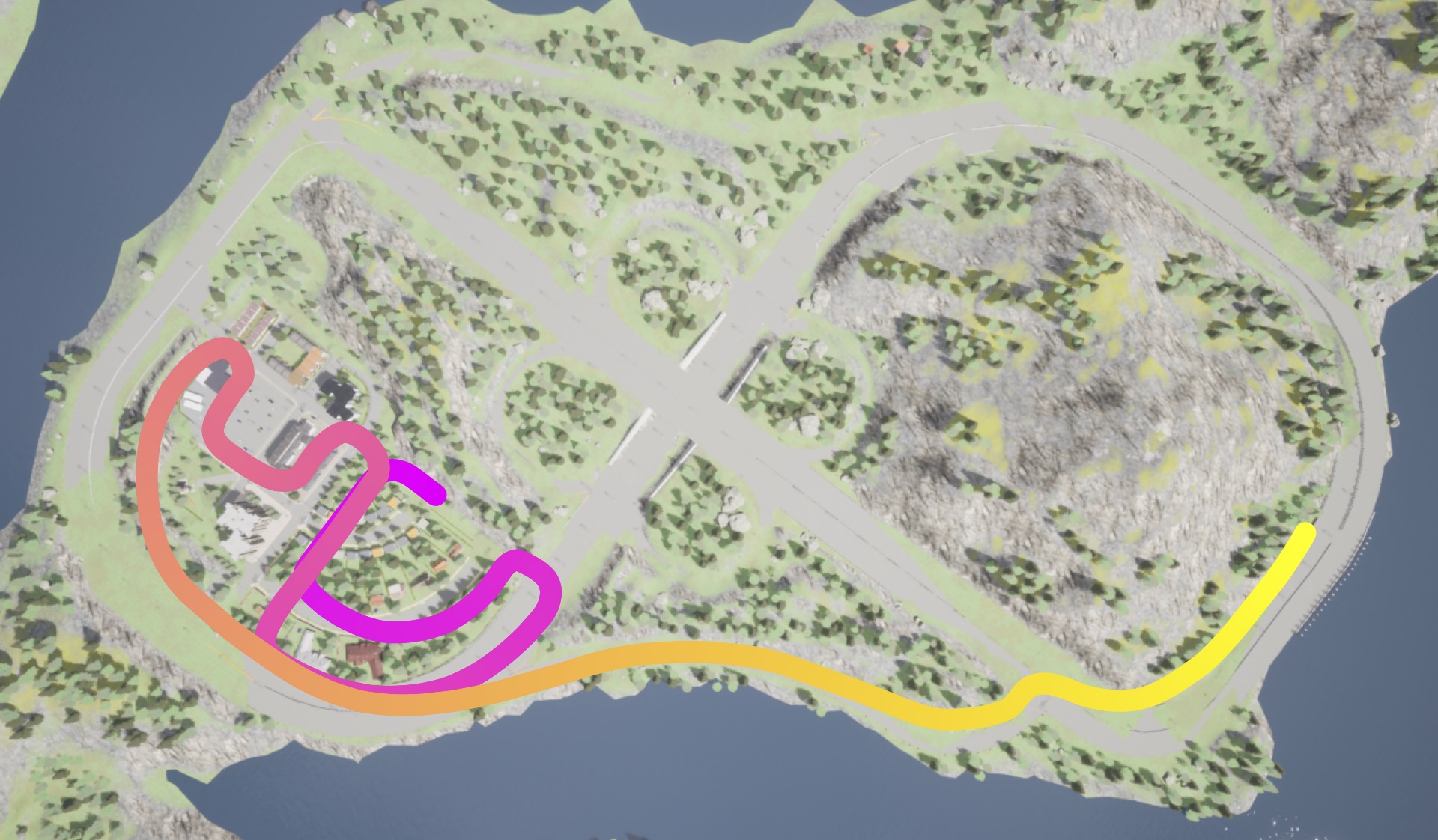}
    \end{subfigure}
    \begin{subfigure}{.24\textwidth}
        \includegraphics[width=\textwidth,height=.7\textwidth]{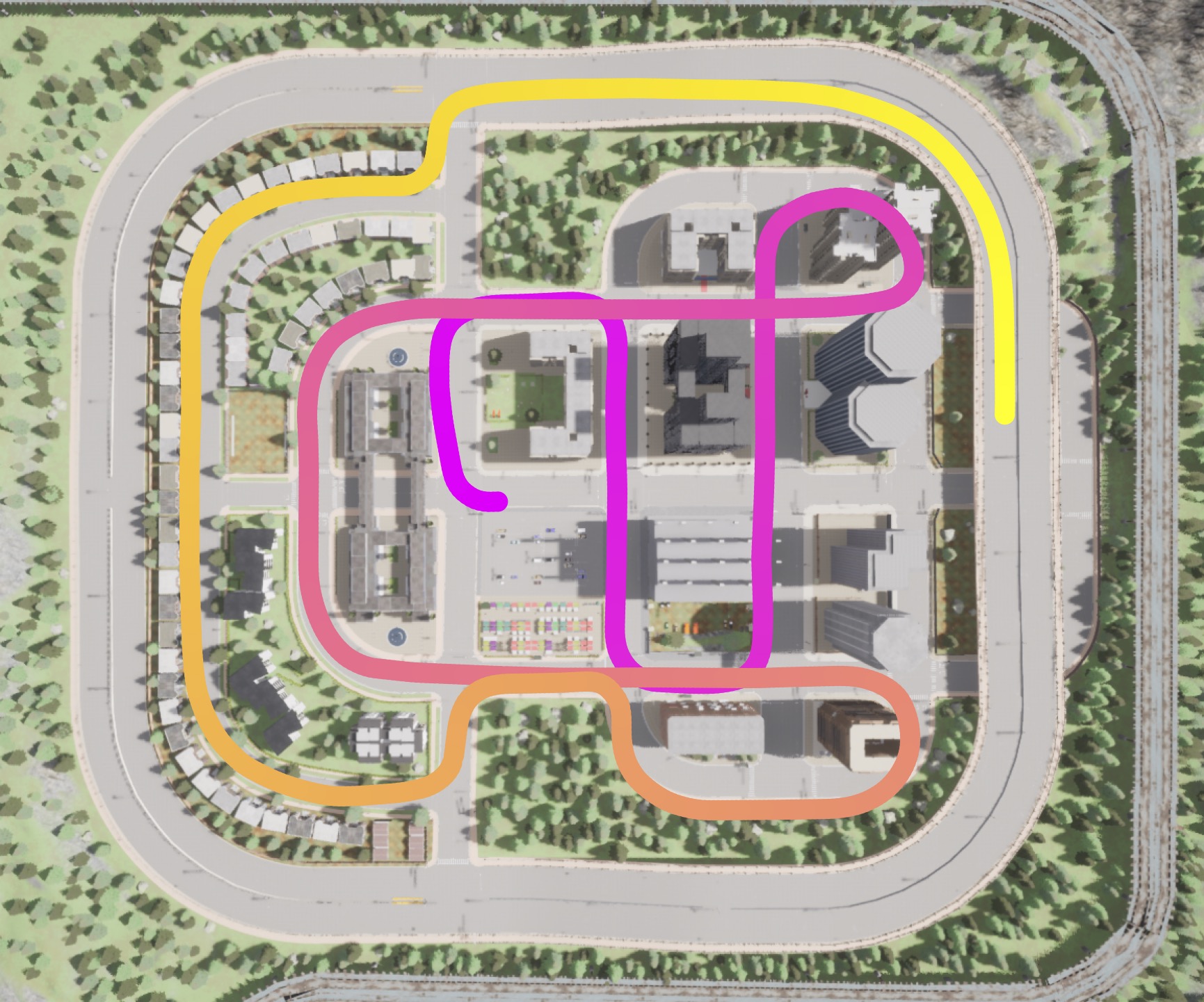}
        \caption*{Town05}
    \end{subfigure} %
    \begin{subfigure}{.24\textwidth}
        \includegraphics[width=\textwidth,height=.7\textwidth]{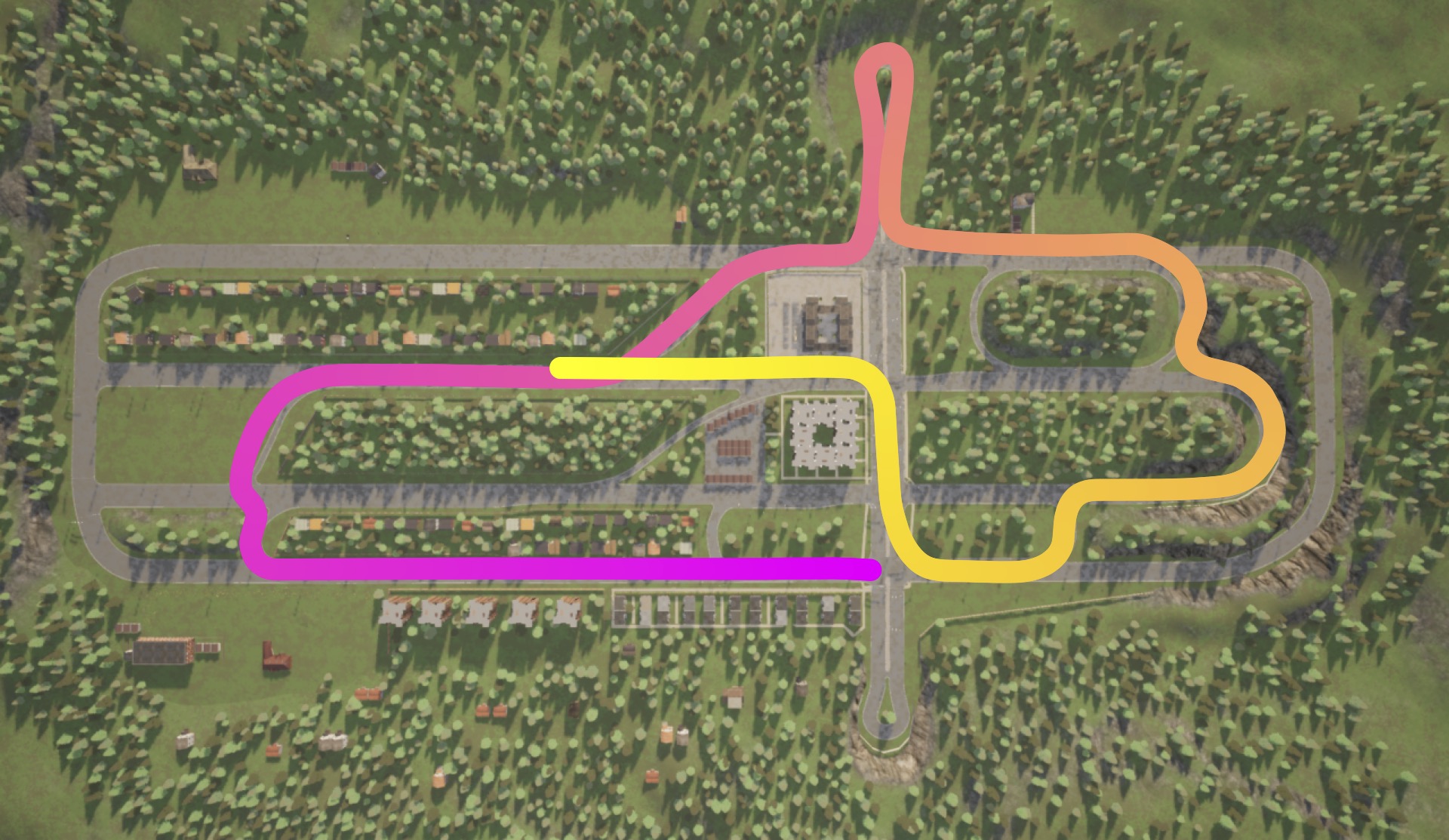}
        \caption*{Town06}
    \end{subfigure} %
    \begin{subfigure}{.24\textwidth}
        \includegraphics[width=\textwidth,height=.7\textwidth]{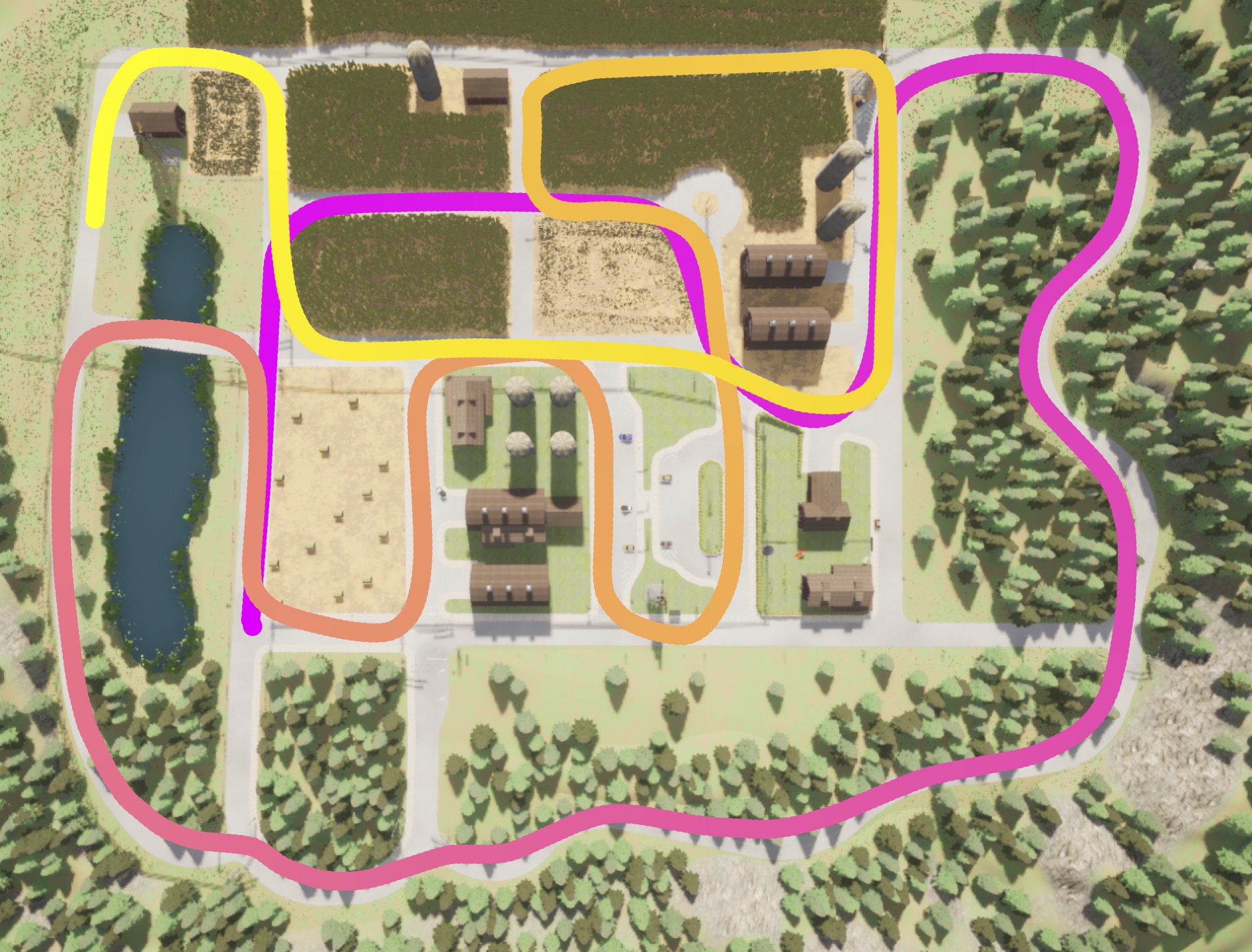}
        \caption*{Town07}
    \end{subfigure} %
    \begin{subfigure}{.24\textwidth}
        \includegraphics[width=\textwidth,height=.7\textwidth]{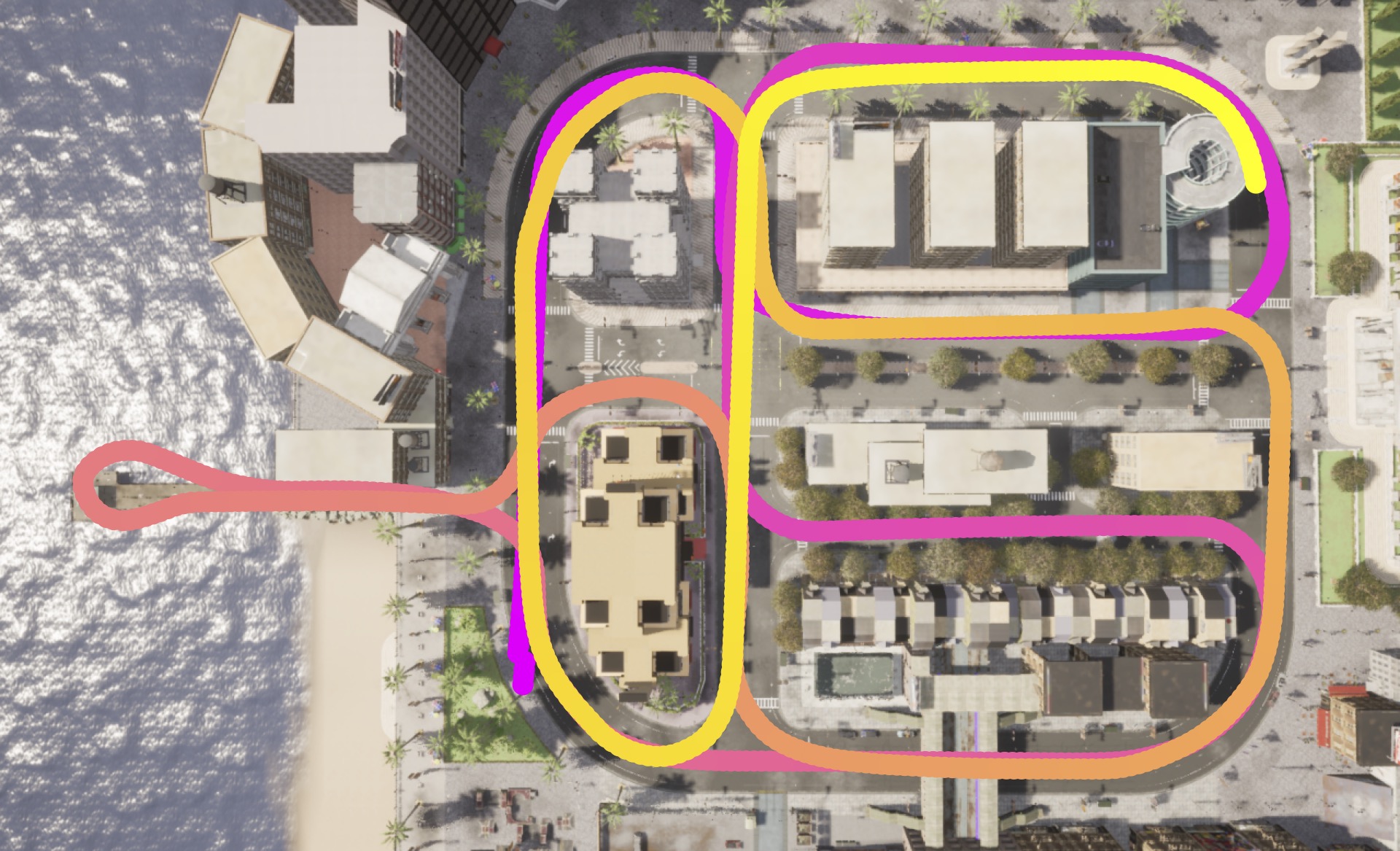}
        \caption*{Town10HD}
    \end{subfigure}
    \caption{Trajectories of the drones exploring the 8 different towns. Start (Pink) to Yellow (End).}
    \label{fig:trajectories}
\end{figure*}
\label{sec:setup}
We adopted a camera sensor setup that leverages the capabilities of the CARLA simulator while ensuring data diversity. The acquisition pipeline involves equipping the UAV with multiple co-registered sensors:

\textit{RGB camera:} It has a resolution of 1920x1080 and enables post-processing effects such as vignette, grain jitter, bloom, auto exposure, lens flare, and depth of field. The vertical Field of View (FoV) is fixed at $90^\circ$, while 
the viewing direction varies based on the selected flying height, with values of $30^\circ$, $60^\circ$, and $90^\circ$ degrees w.r.t the horizontal axis for altitudes of 20, 50, and 80 meters,  respectively. The color images are saved in JPEG format.

\textit{Depth camera:} The depth camera has the same FoV and resolution as the RGB one. It has a $1 km$ maximum range. %
Depth images are saved in PNG with the format of \cite{ros2016synthia}.

\textit{Lidar sensor:} It is a 64-channel sensor with a $360^\circ$ FoV, operating at bird view ($-89^\circ$ to $-49^\circ$ \wrt the horizontal) and collecting $\sim$100k points for each acquisition. 
The Lidar data is provided for the height of $80m$ and has a maximum range of $100m$. This results in a field of view at road level of $74^{\circ}$, \ie, 60m of distance from the position perpendicular to the drone. The remaining channels aid in the identification of tall objects (such as buildings or trees). The horizontal resolution is $0.230^\circ$ and the vertical $0.625^\circ$. %

The Unreal Engine utilized in \dname incorporates a physics engine, specifically based on NVIDIA PhysX, which simulates the movement of vehicles, pedestrians, and sensors. This enables the generation of realistic trajectories and motion blur effects due to drone movements. The velocity vector of the drone is taken into account to calculate the motion blur accurately, providing a more realistic visual representation.
The sensor undergoes rotations at a frequency of 25 Hz, which matches the frequency of the cameras used in the dataset. Consequently, each color/depth frame has a corresponding full $360^{\circ}$ lidar scan.

\subsection{Data specifications
}

The data were acquired by simulating the flight of a drone across the 8 virtual towns. Figure \ref{fig:trajectories} shows the trajectory followed inside each of the towns. Each trajectory has a length of about 2-3km which corresponds to a 2 minutes flight ($\sim20m/s$). Sensor data is recorded at 25 fps, for a total of 3000 frames for each sensor.
To extract testing sequences with a class distribution as close as possible to the training data, while still avoiding too close frames in the train and test sets, we opted to extract 5 equispaced sub-sequences of 100 samples (4 seconds each, 20 seconds in total) from each of the rendered trajectories. This corresponds to a total of $8*3*500 = 12k$ test samples.

In order to simulate drone trajectories for road surveillance, \dname has been designed to mimic real-world scenarios where drones are deployed for applications like monitoring road traffic volume or detecting accidents. In such cases, the drone's viewpoint can vary as it adjusts its altitude to capture different perspectives. 
For this reason, to enhance the dataset's robustness and generalization capabilities, \dname records data from various heights and view angles.
Moreover, we provide ground truth (GT) annotations in both the form of pixel-level semantic maps and 3D bounding boxes with unique identifiers (IDs) for all actors in the scene, including vehicles and pedestrians, at each temporal instant. These annotations enable researchers to perform comprehensive analyses of both semantic segmentation and 2D or 3D object detection methods, enhancing the development of advanced algorithms and systems for UAV-based vision applications.

\section{Benchmark and Experiments}

We start by reporting some benchmark results of various architectures on our dataset. For consistency and reproducibility, all of the models used the official implementation by the torchvision library\footnote{Pytorch segmentation models available \href{https://pytorch.org/vision/stable/models.html\#semantic-segmentation}{here} and the object detection ones \href{https://pytorch.org/vision/stable/models.html\#object-detection}{here}. Accessed 10-July-2023.}.

In particular, we employ the widely used
\textbf{Deeplab-V3} \cite{chen2017rethinking} network with both the ResNet50 and MobileNetV2 backbones for the Semantic Segmentation task. The choice of the two backbones follows the idea of having both a highly-performing backbone for server-side computation and a lightweight one that could be used onboard. 
For the Object Detection task we used \textbf{FasterRCNN} \cite{ren2015faster} and \textbf{RetinaNet} \cite{lin2017focal}. 
The overall performance and computational cost in terms of MACs (Multiply-Add Cumulation) of the architectures are reported in Table \ref{tab:backbones}.
The models were trained for 60k iterations with a batch size of 2. The learning rate was set to 2.5e-4, and a cosine annealing scheduler with a linear warmup for 2000 steps was employed. The semantic segmentation task considers 28 classes, while object detection includes 8 classes, that consist of the %
the moving objects (\ie, the vehicles and pedestrians).
For the object detection task, the rider and motorcycle classes are combined to form the class motorcyclist, while the classes rider and bicycle are merged into the class bicyclist.

\begin{table}[t]
\centering
\begin{tabular}{|c|llll|}
\hline
\diagbox[]{Train}{Test} & All & 20m & 50m & 80m \\
\hline
All & \textbf{61.1} & 63.0 & 60.6 & 56.2 \\ %
20m & 48.4 & \textbf{65.1} & 46.0 & 28.1 \\ %
50m & 50.7 & 41.3 & \textbf{61.4} & 52.5 \\ %
80m & 42.9 & 25.8 & 51.8 & \textbf{57.9} \\
\hline
\end{tabular}
\caption{mean Intersection over Union (mIoU) in the semantic segmentation task with data at different altitudes.}
\label{tab:altitudes}
\end{table}

\begin{table}[t]
\setlength{\tabcolsep}{3pt}
\centering
\resizebox{\columnwidth}{!}{
\begin{tabular}{|c|ccccccccc|}
\hline
\diagbox[]{Train}{Test} & all & t01 & t02 & t03 & t04 & t05 & t06 & t07 & t10 \\
\hline
all & \textbf{61.1} & 48 & 44 & 58.2 & 47.1 & 52.7 & 41.7 & 43.1 & 44 \\
t01 & 21.2 & \textbf{55.8} & 27.9 & 16.1 & 20.3 & 17.4 & 15.8 & 22 & 6.7 \\
t02 & 16.8 & 25.4 & \textbf{57.5} & 12 & 13.2 & 12.8 & 10.8 & 12.6 & 7.2 \\
t03 & 28.3 & 15.6 & 19.9 & \textbf{64.3} & 23.4 & 26.3 & 24.1 & 15 & 15 \\
t04 & 24.5 & 17 & 15.3 & 21.2 & \textbf{54.7} & 24.8 & 23 & 18.3 & 11.4 \\
t05 & 25.2 & 14.2 & 14.6 & 25.7 & 22.8 & \textbf{58} & 25.3 & 13.8 & 10.5 \\
t06 & 16 & 10.5 & 9.6 & 15.1 & 20.2 & 17 & \textbf{48.7} & 12.9 & 8.5 \\
t07 & 18 & 15.3 & 12.6 & 12.5 & 20.8 & 16.3 & 19.6 & \textbf{53.1} & 4.3 \\
t10 & 21.6 & 12.3 & 13.9 & 16.6 & 14.2 & 17.2 & 15.8 & 8 & \textbf{53.3} \\
\hline 
\end{tabular}}
\caption{mIoU for the semantic segmentation task across different towns (t=town). %
}
\label{tab:towns}
\end{table}

\subsection*{Evaluation at  different flying altitudes}

We performed four different trainings on the model, one for each of the three flying altitudes (20m, 50m, and 80m) and one considering all heights together.  
In Table \ref{tab:altitudes}, we provide a comprehensive overview of the model's performance at different test altitudes.
As expected, the model trained on the entire dataset demonstrates the highest overall accuracy when testing on data at all altitudes ($61.1\%$), suggesting that incorporating various heights during training facilitates improved performance across different altitudes. Moreover, the data in the table highlights that altitudes closer to each other exhibit similar performances (with the best performances when training and testing at the same altitude), while the model's generalization tends to decrease as the difference in altitude between training and testing data increases. 
Furthermore, it can be noticed that the training at the lowest altitude displays a significant drop in accuracy when tested at the highest altitude, achieving a mere $28.1\%$. As expected, since at higher altitudes the objects appear smaller and thus harder to be recognized, the model seems to struggle to effectively generalize to this setting, especially when trained at lower altitudes. These observations highlight the importance of considering the interplay between different altitudes. 

\begin{table}[t]
\setlength{\tabcolsep}{2pt}
\centering
\resizebox{\columnwidth}{!}{
\begin{tabular}{cccccc}
\hline
\textbf{Model} & \textbf{Backbone} & \begin{tabular}[c]{@{}c@{}}\textbf{GMAC }\cite{ptflops}\\ @(1080x1920)\end{tabular} & \textbf{mIoU} &  \textbf{mAP@50} & \textbf{mAP@75}\\
\midrule
\multirow{2}{*}{DeepLabV3 \cite{chen2017rethinking}} & MNv3 & 78.9 & 61.1 & - & -\\
& RN50 & 1297.22 & 72.0 & - & - \\
\midrule
Faster R-CNN \cite{ren2015faster} & MNv3  & 8.35 & - & 31.1 & 15.7 \\
RetinaNet \cite{lin2017focal} & RN50 & 207.94 & - & 36.2 & 30.4\\
\bottomrule
\end{tabular}}
\caption{Comparison of different models over Semantic Segmentation and Object Detection tasks. We also report the computational complexity in terms of MACs. %
Note for the reader: FLOPs $\simeq$ $2*$MACs, RN50=ResNet50, MNv3=MobileNetV3 Large.\\}
\label{tab:backbones}
\centering
\begin{tabular}{ccc}
\hline
\textbf{Data} & \begin{tabular}[c]{@{}c@{}}\textbf{GMAC}\cite{ptflops}\\ @(1080x1920)\end{tabular} & \textbf{mIoU} \\

\midrule
RGB & \multirow{3}{*}{78.9} & 61.1 \\
D &  & 59.1 \\
RGB+D (early) &  & 60.7 \\
RGB+D (late) & 161.05 & \textbf{64.2} \\ 
\bottomrule
\end{tabular}
\caption{Comparison of the training over different modalities.}
\label{tab:mm}
\end{table}

\begin{table*}[t]
\centering
\def\arraystretch{1.1}
\begin{tabular}{cccccccccc}
\toprule
\multirow{2}{*}{Setting} & \multicolumn{2}{c}{Aeroscapes} & \multirow{2}{*}{ICG Drone} & \multicolumn{2}{c}{UAVid} & \multicolumn{2}{c}{UDD5} & \multicolumn{2}{c}{UDD6} \\ \cline{2-3} \cline{5-10}
 & train & val &  & train & val & train & val & train & val \\
\midrule
Oracle & 90.9 & 71.2 & 94.1 & 87.7 & 70.4 & 97.1 & 86.4 & 97.3 & 84.8\\
\midrule
\dname (w/o Resize, w AllClasses) & 22.0 & 27.5 & 15.6 & 32.9 & 33.7 & 27.2 & 26.7 & 26.8 & 26.2\\
\dname (w Resize, w AllClasses) & 22.0 & 27.5 & 16.8 & 36.3 & 35.6 & 30.9 & 30.0 & 30.2 & 29.7\\
\dname (w Resize, w/o AllClasses) & 24.6 & 33.1 & 16.8 & 51.3 & 53.6 &  58.3 & 57.1 & 57.2 & 56.4\\
\bottomrule
\end{tabular}
\caption{Performance of models trained on \dname and tested on  other semantic segmentation datasets. Refer to Table \ref{tab:mapping} for the class re-mapping. In the latter tests only the valid classes for each specific dataset are considered. %
}
\label{tab:adapt}
\end{table*}
\newcommand{\imsize}{.19\textwidth}
\begin{figure*}[t]
    \centering
    \begin{subfigure}{10px}
        \rotatebox{90}{\hspace{1em} Pred \hspace{3em} GT}
    \end{subfigure}%
    \begin{subfigure}{\imsize}
        \begin{subfigure}{\textwidth}
            \caption*{Aeroscapes}
            \includegraphics[width=\textwidth,trim=0 0 0 50,clip]{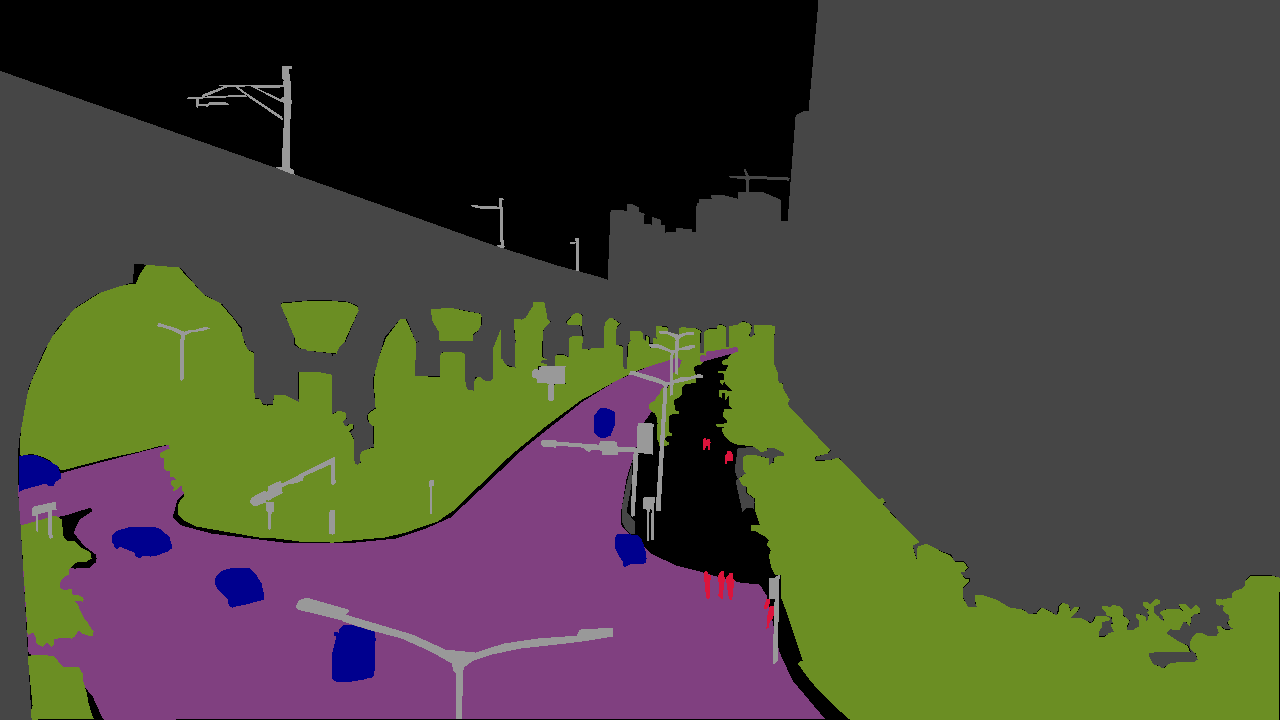}
        \end{subfigure}
        \begin{subfigure}{\textwidth}
            \includegraphics[width=\textwidth,trim=0 0 0 50,clip]{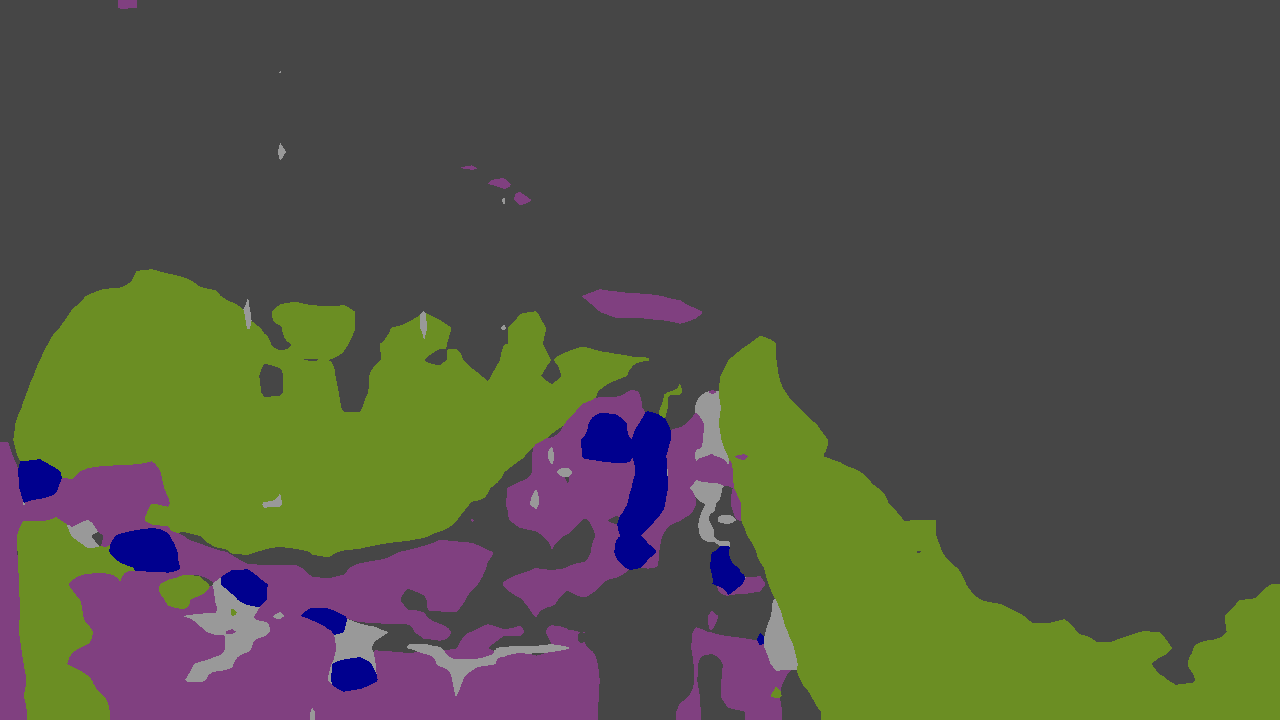}
        \end{subfigure}
    \end{subfigure}
    \begin{subfigure}{\imsize}
        \begin{subfigure}{\textwidth}
        \caption*{ICG Drone}
            \includegraphics[width=\textwidth,trim=0 270 0 0,clip]{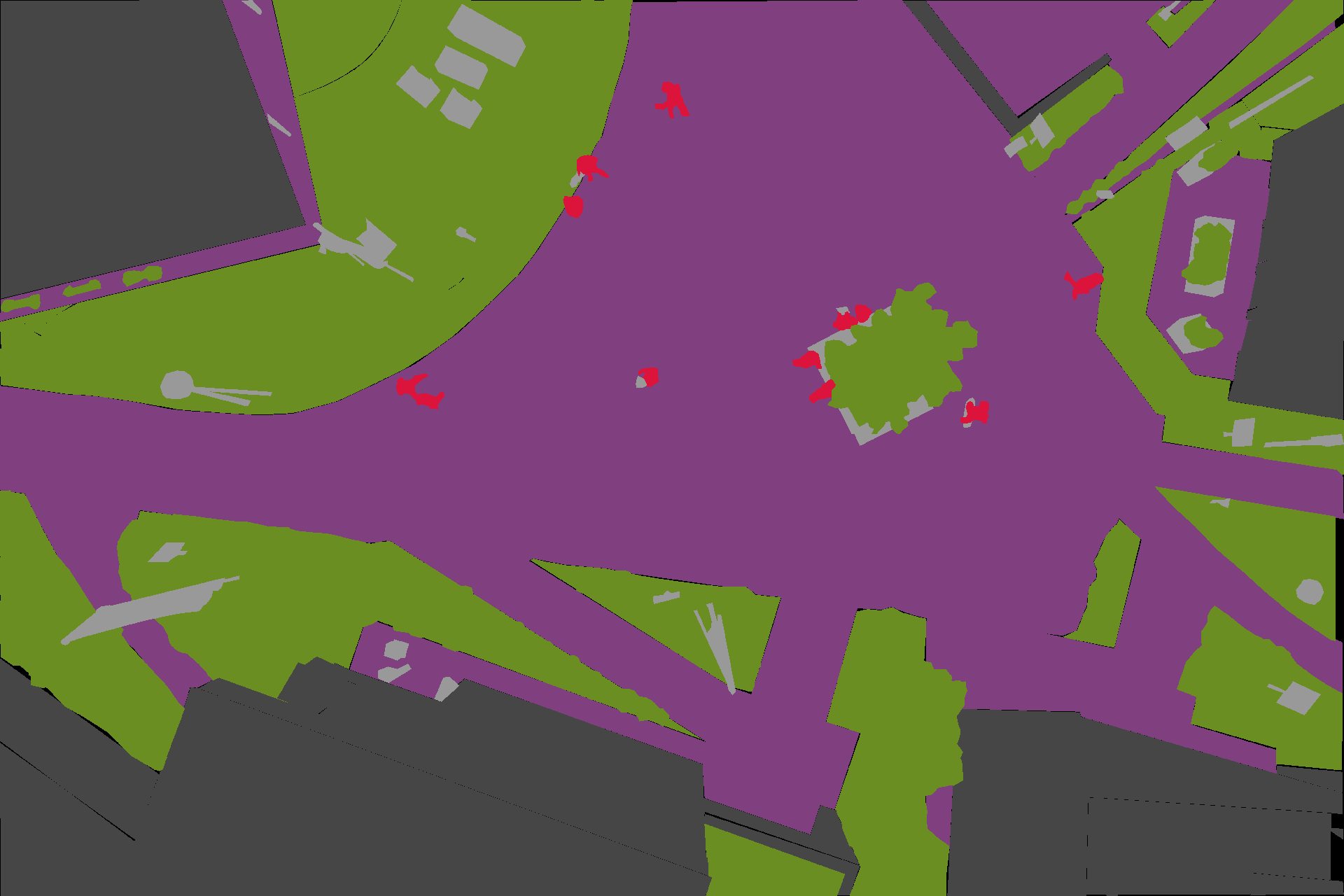}
        \end{subfigure}
        \begin{subfigure}{\textwidth}
            \includegraphics[width=\textwidth,trim=0 270 0 0,clip]{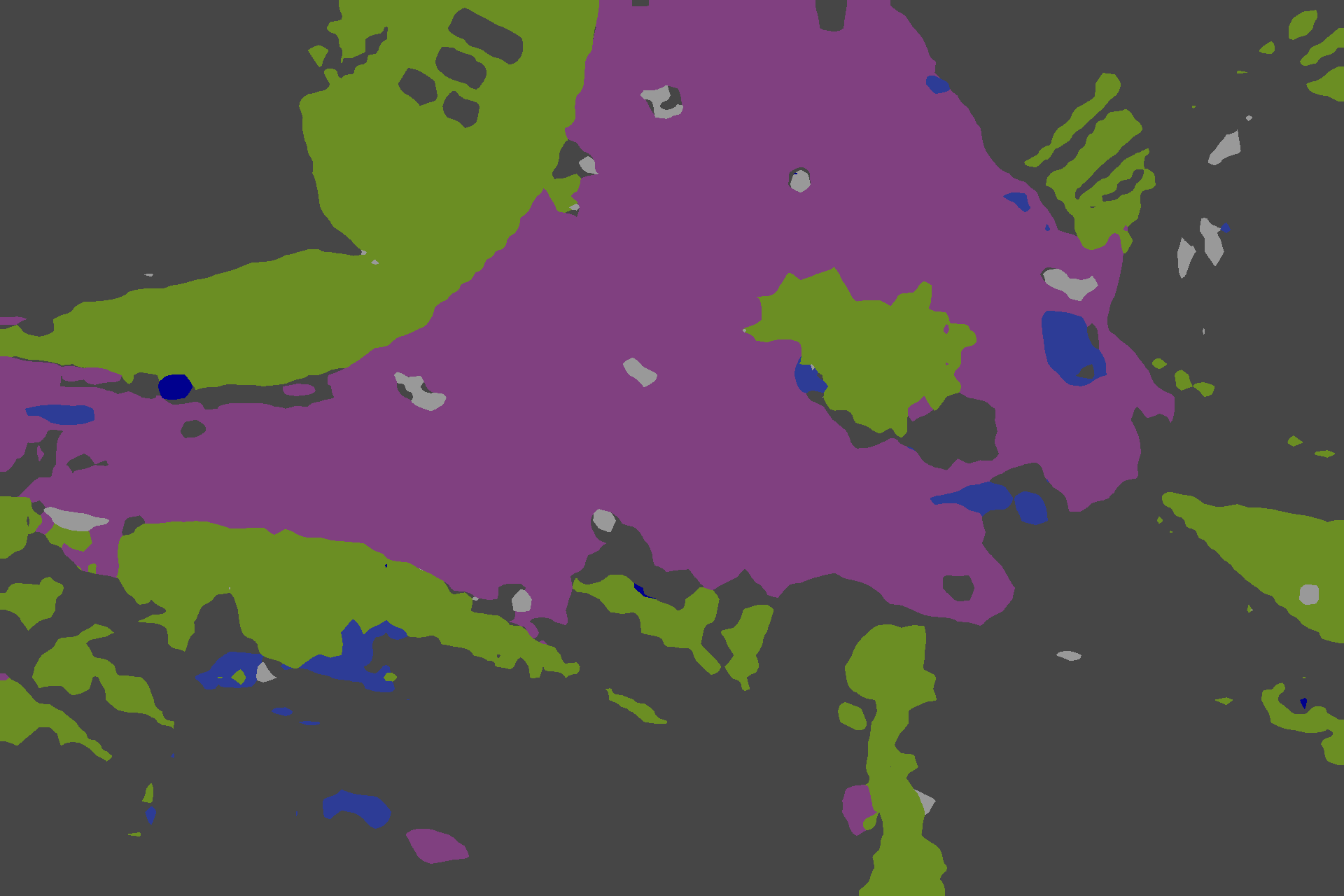}
        \end{subfigure}
    \end{subfigure}
    \begin{subfigure}{\imsize}
        \begin{subfigure}{\textwidth}
        \caption*{UAVid}
            \includegraphics[width=\textwidth,trim=0 15 0 15,clip]{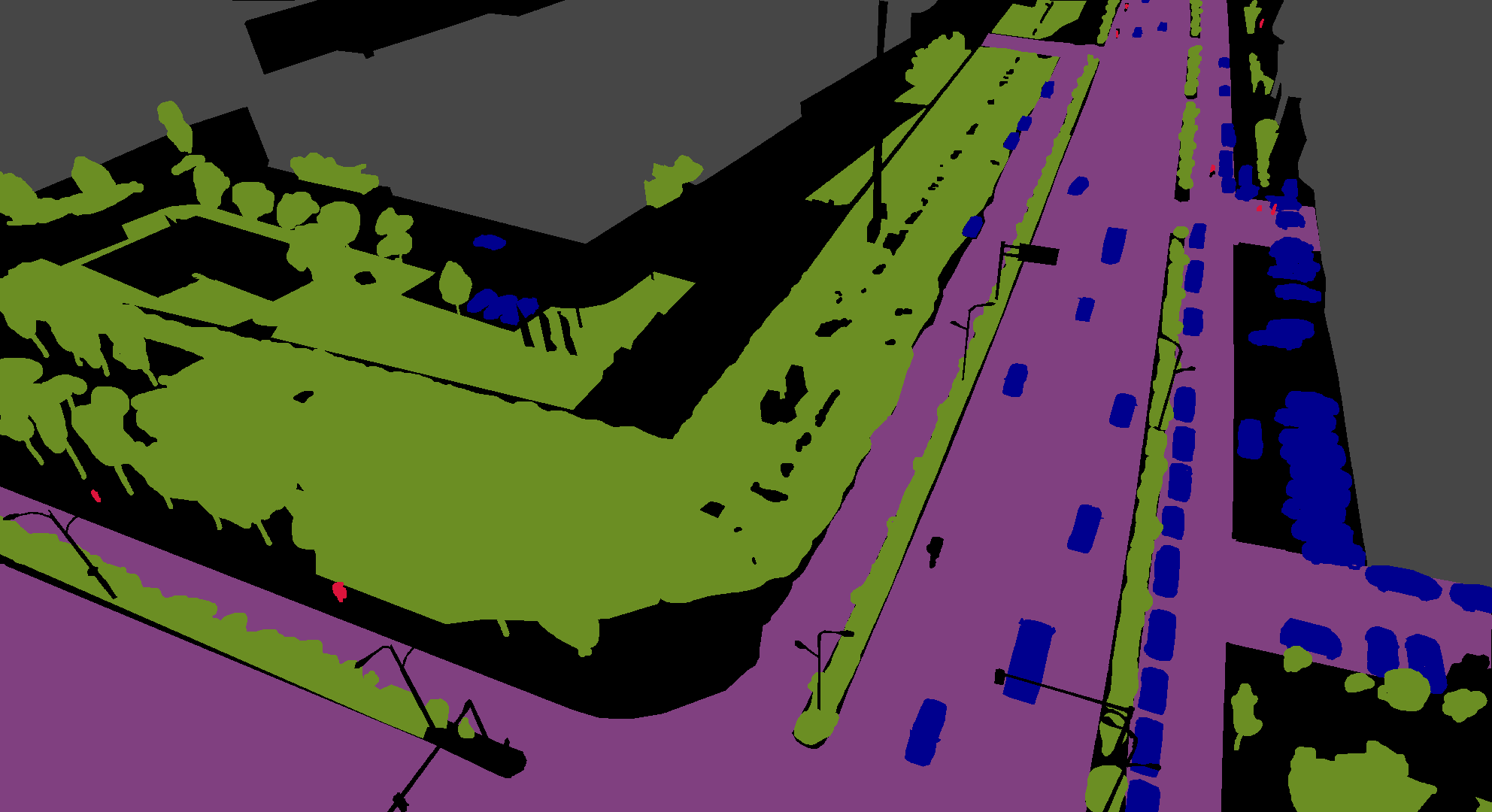}
        \end{subfigure}
        \begin{subfigure}{\textwidth}
            \includegraphics[width=\textwidth,trim=0 15 0 15,clip]{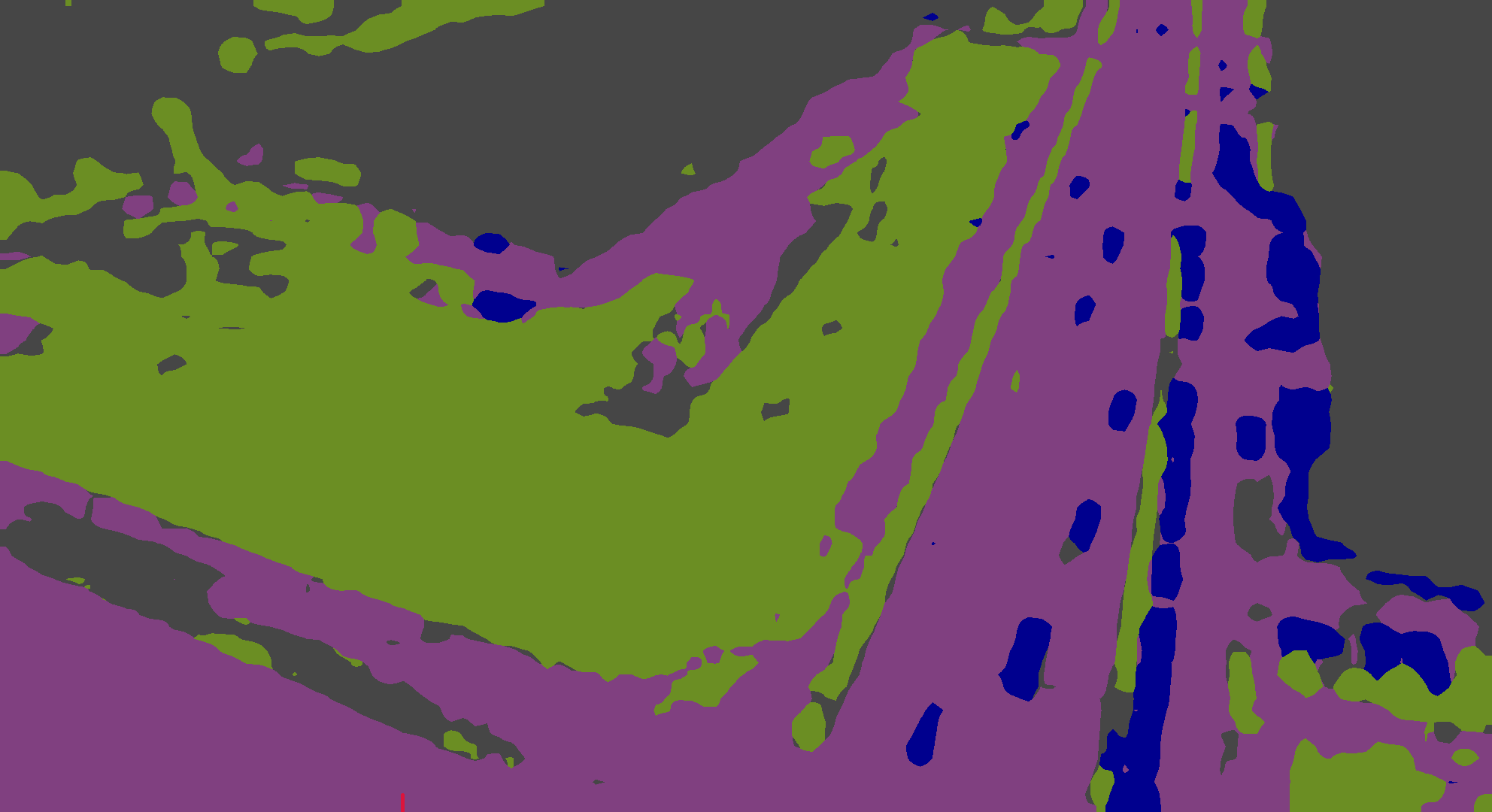}
        \end{subfigure}
    \end{subfigure}
    \begin{subfigure}{\imsize}
        \begin{subfigure}{\textwidth}
        \caption*{UDD5}
            \includegraphics[width=\textwidth]{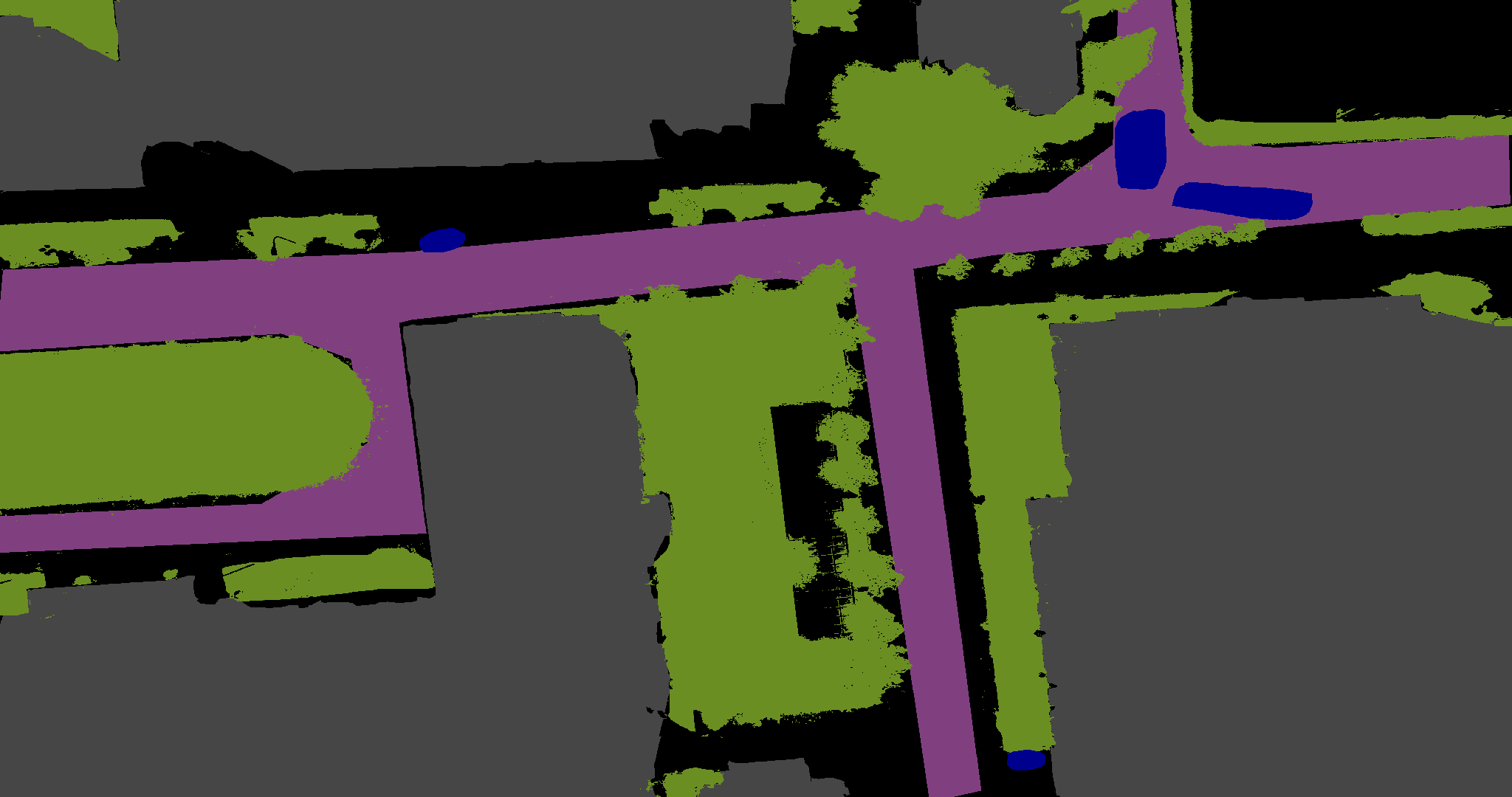}
        \end{subfigure}
        \begin{subfigure}{\textwidth}
            \includegraphics[width=\textwidth]{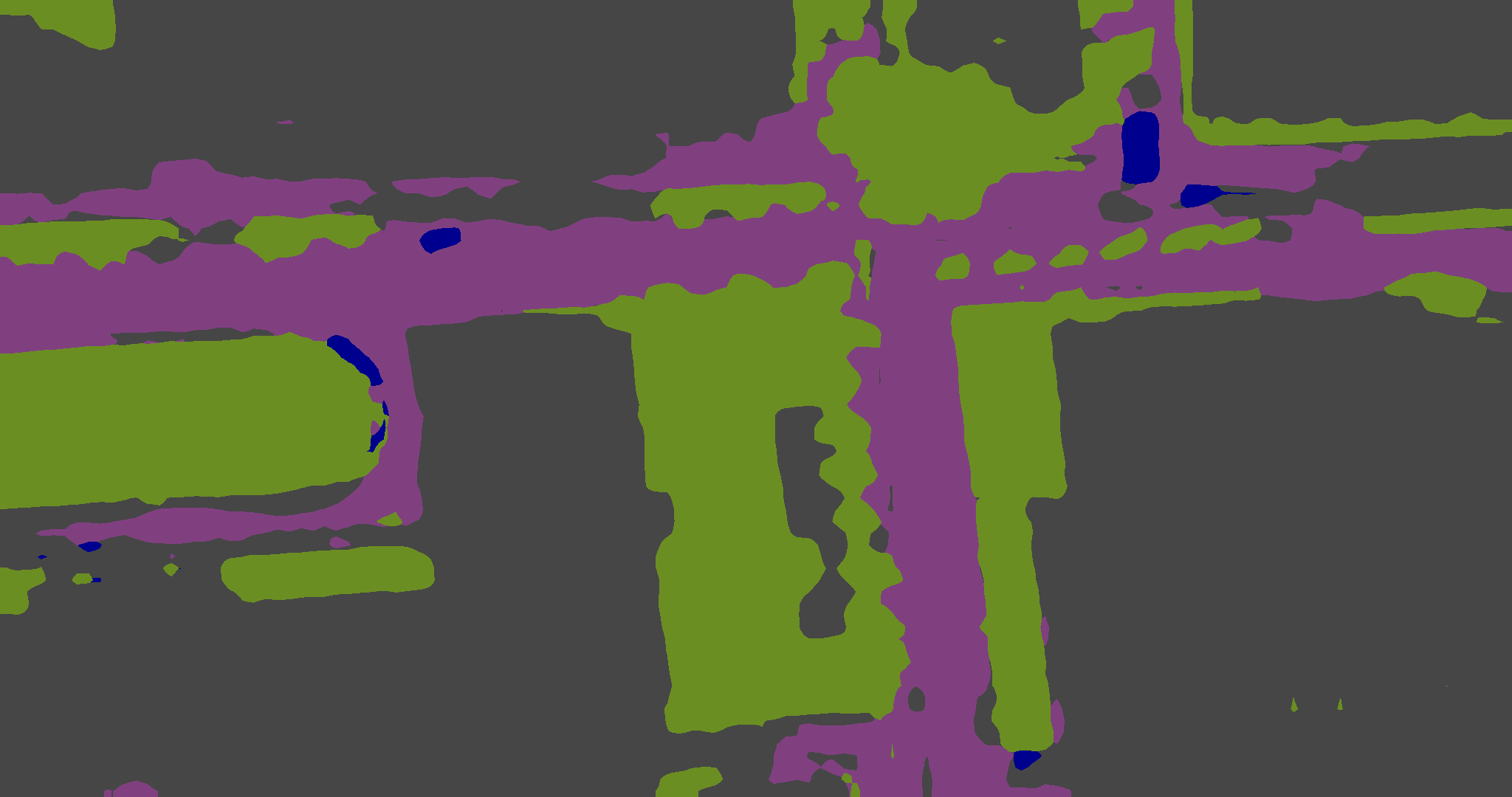}
        \end{subfigure}
    \end{subfigure}
    \begin{subfigure}{\imsize}
        \begin{subfigure}{\textwidth}
        \caption*{UDD6}
            \includegraphics[width=\textwidth]{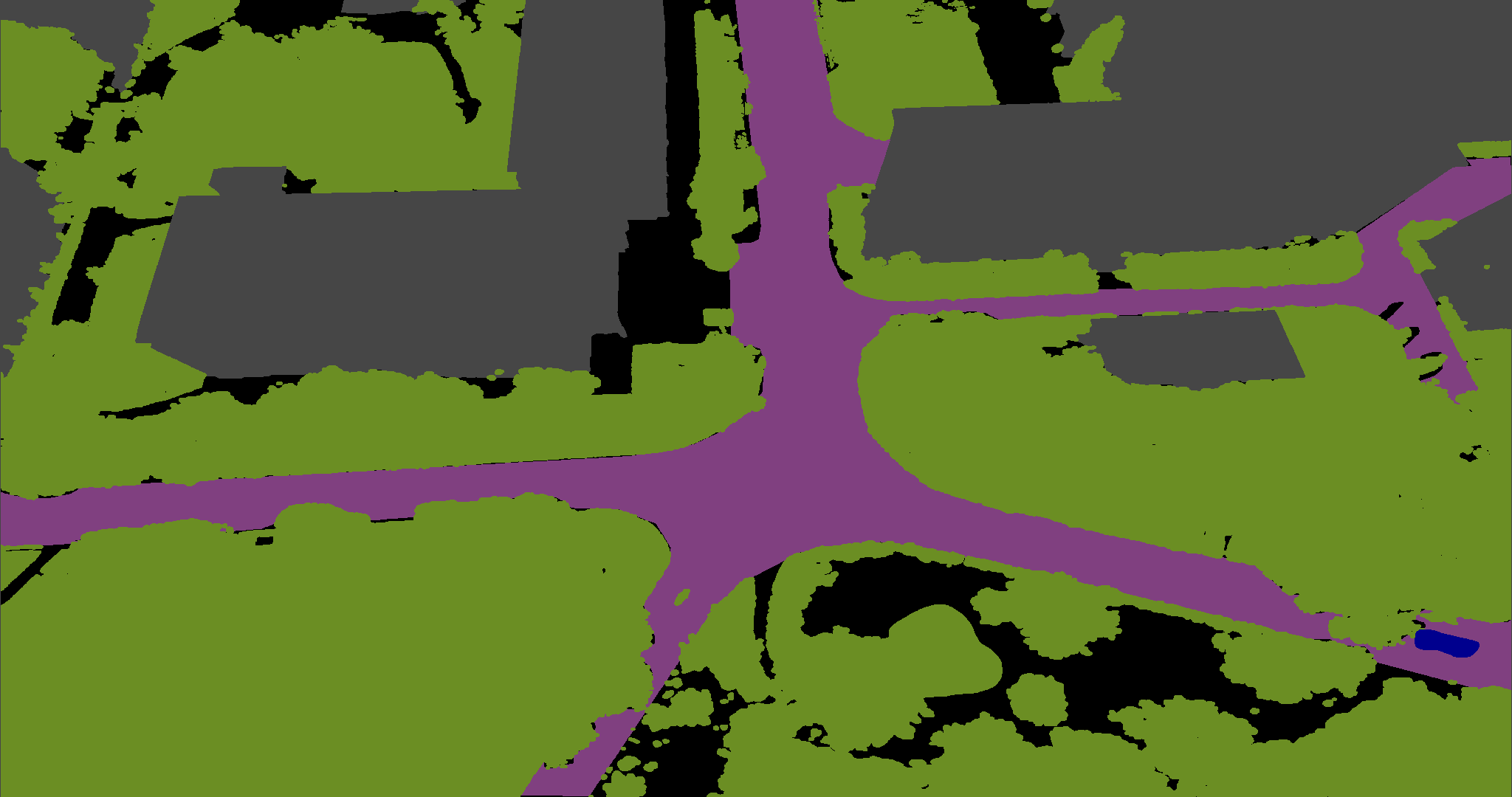}
        \end{subfigure}
        \begin{subfigure}{\textwidth}
            \includegraphics[width=\textwidth]{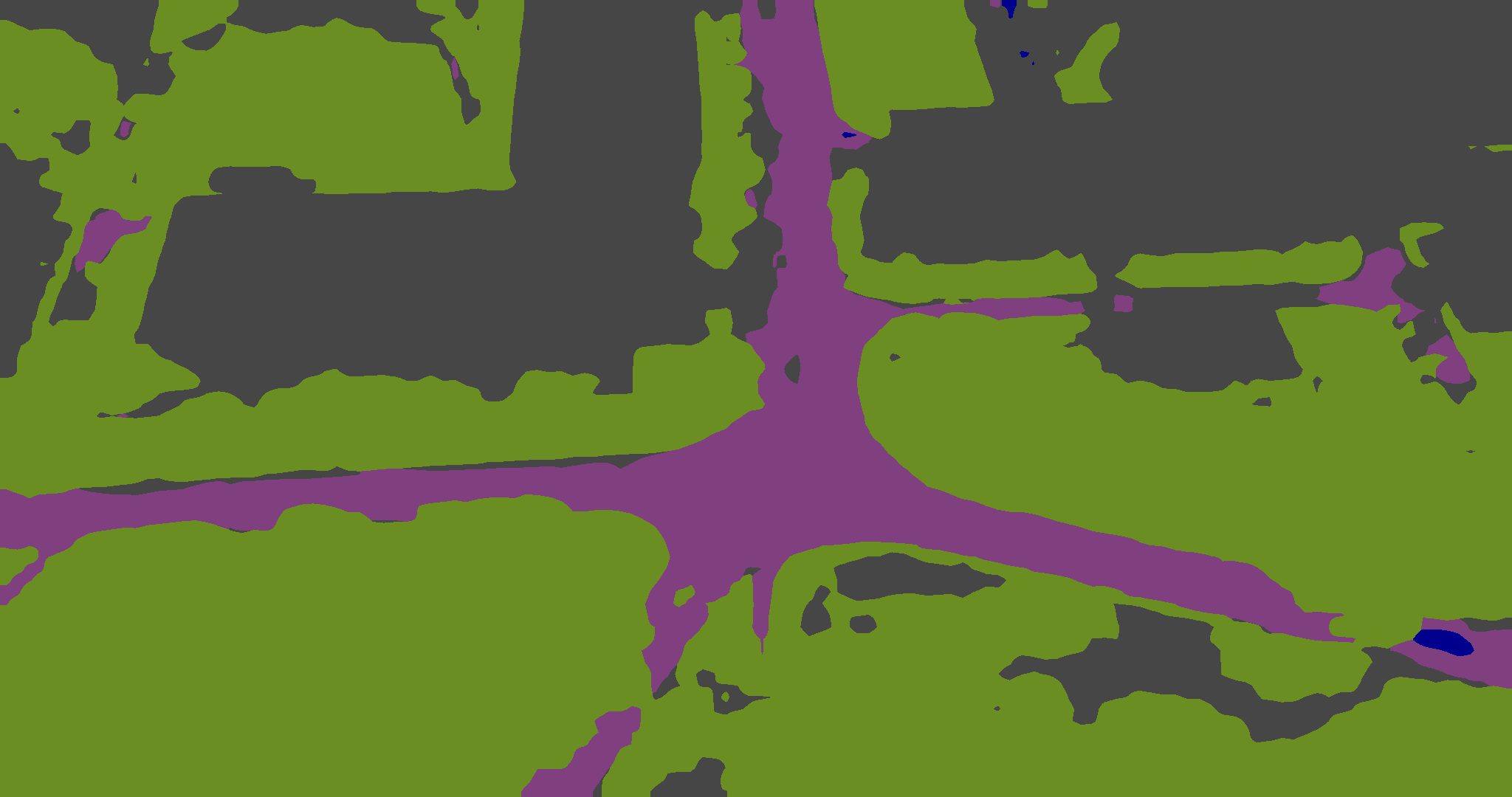}
        \end{subfigure}
    \end{subfigure} %
    \begin{subfigure}{10px}
        \rotatebox{90}{\hspace{1em} Pred \hspace{3em} GT}
    \end{subfigure}%
    \begin{subfigure}{\imsize}
        \begin{subfigure}{\textwidth}
            \includegraphics[width=\textwidth,trim=0 0 0 50,clip]{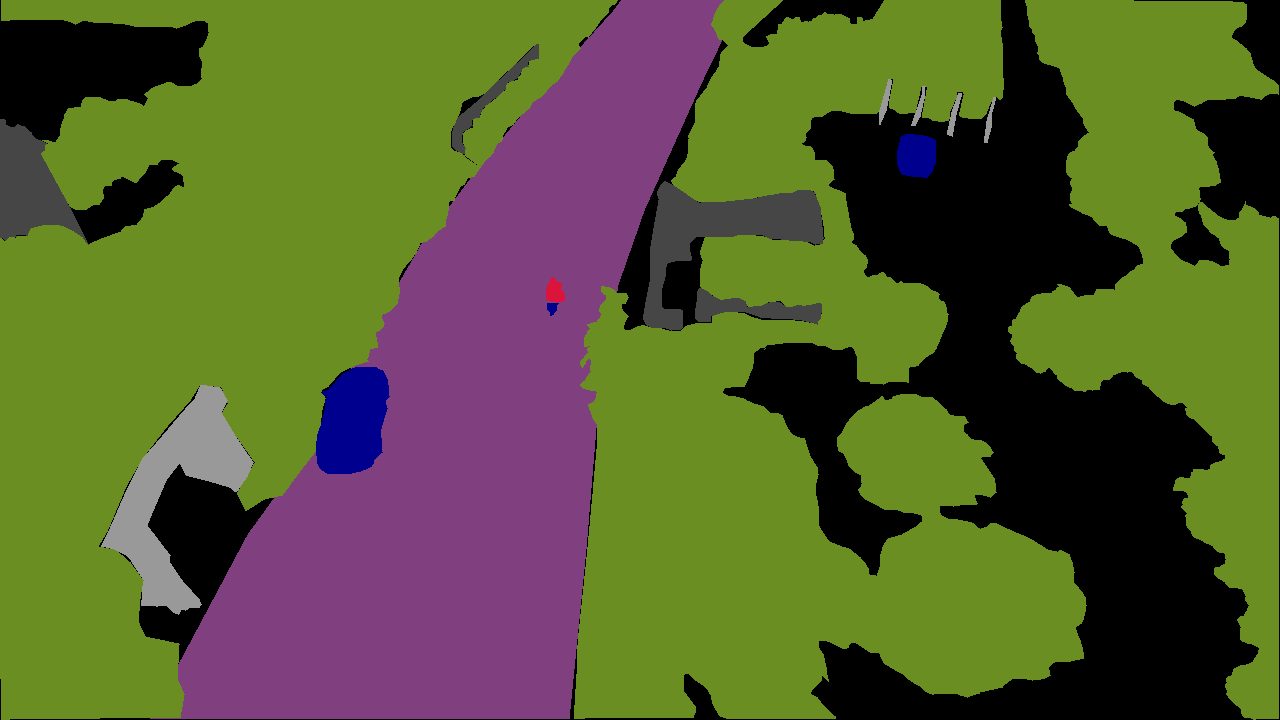}
        \end{subfigure}
        \begin{subfigure}{\textwidth}
            \includegraphics[width=\textwidth,trim=0 0 0 50,clip]{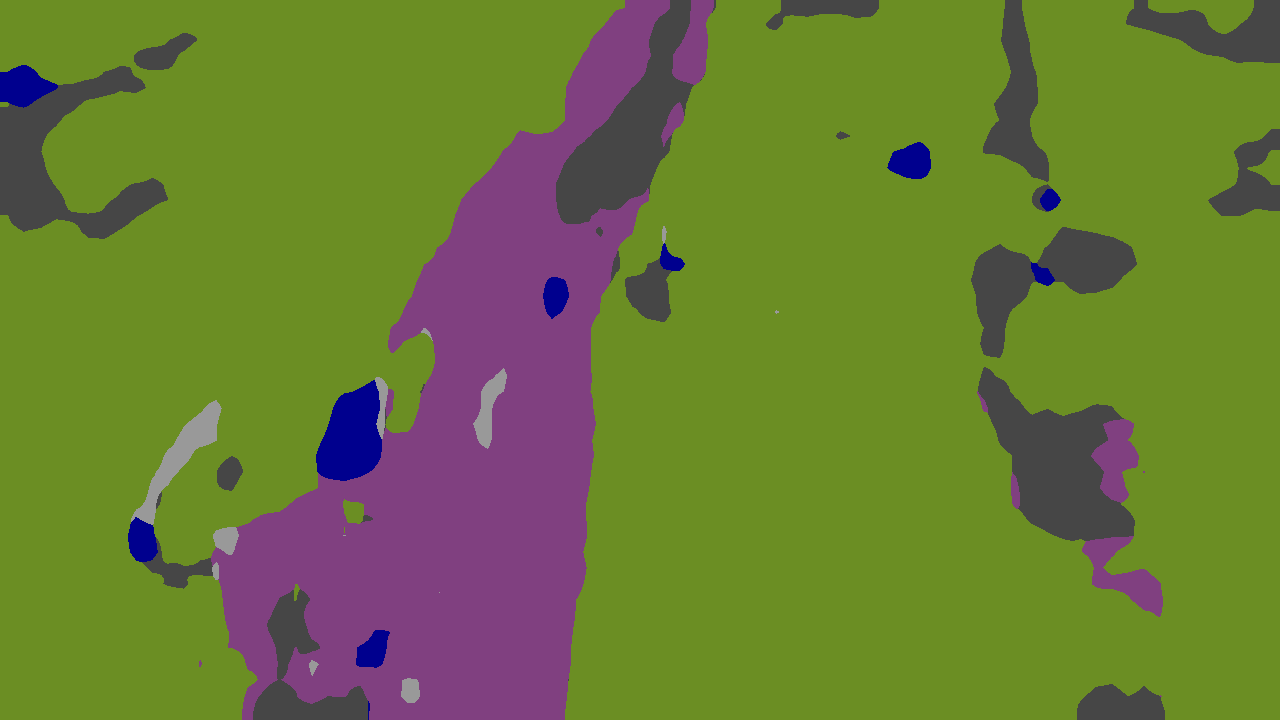}
        \end{subfigure}
    \end{subfigure}
    \begin{subfigure}{\imsize}
        \begin{subfigure}{\textwidth}
            \includegraphics[width=\textwidth,trim=0 270 0 0,clip]{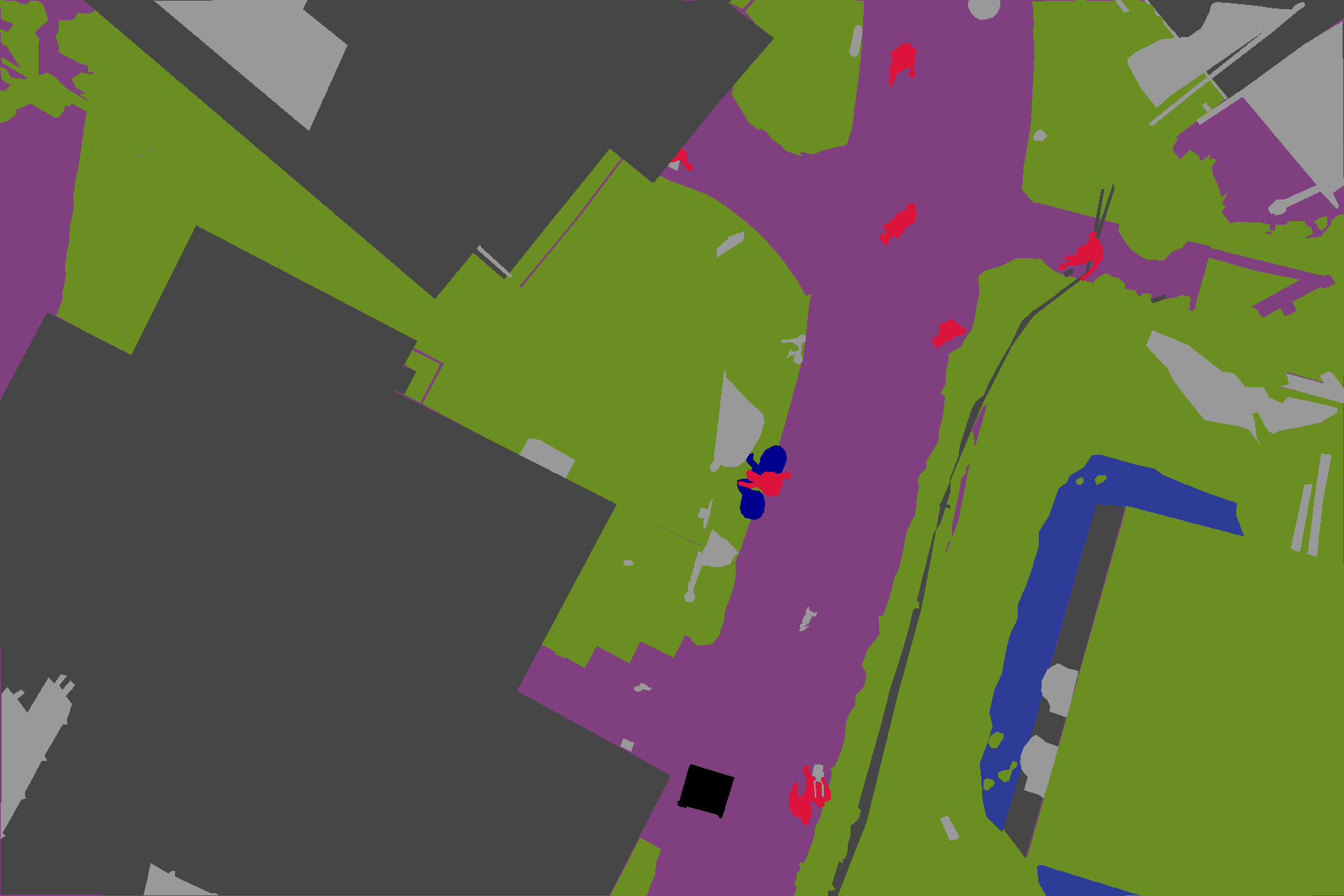}
        \end{subfigure}
        \begin{subfigure}{\textwidth}
            \includegraphics[width=\textwidth,trim=0 270 0 0,clip]{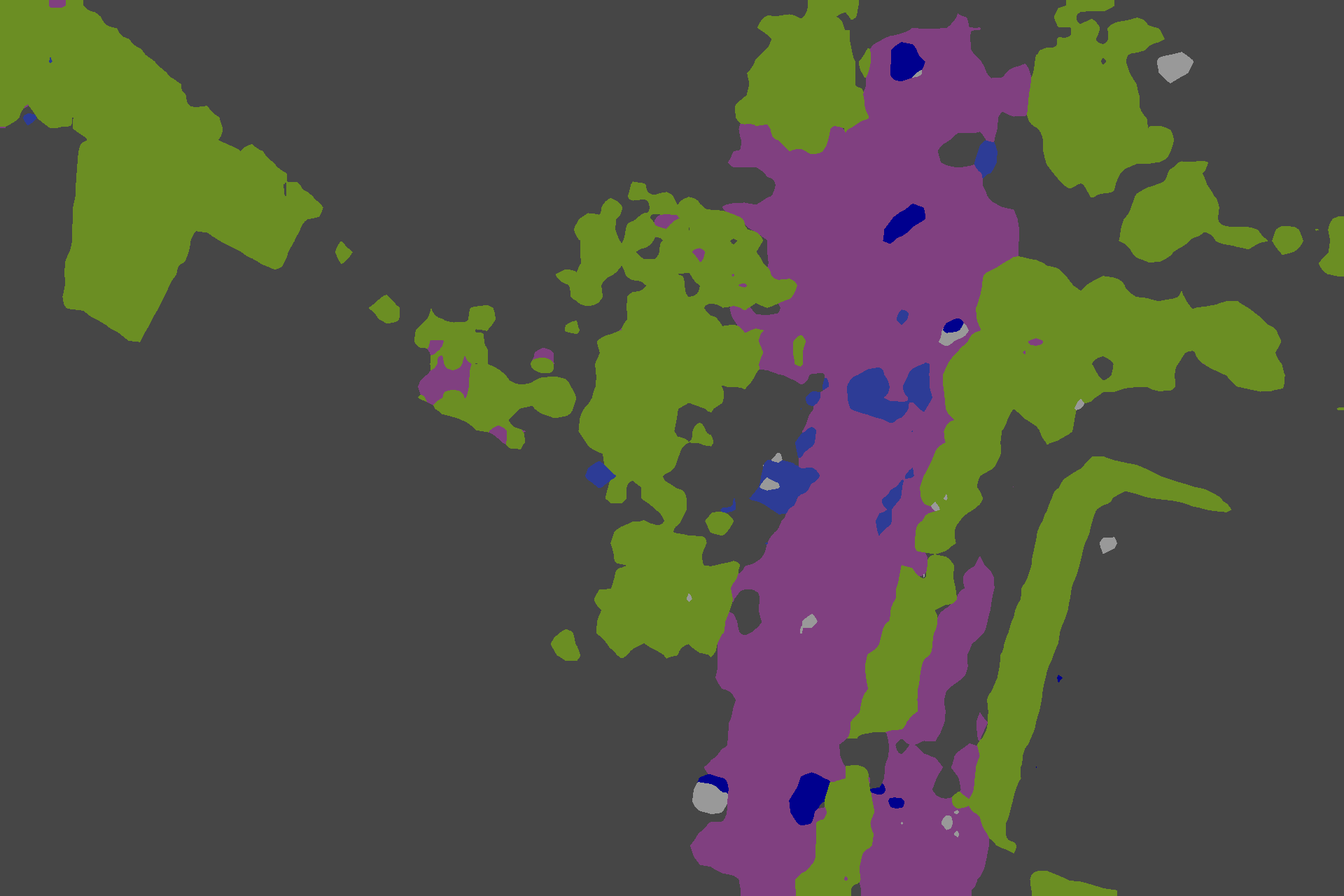}
        \end{subfigure}
    \end{subfigure}
    \begin{subfigure}{\imsize}
        \begin{subfigure}{\textwidth}
            \includegraphics[width=\textwidth,trim=0 15 0 15,clip]{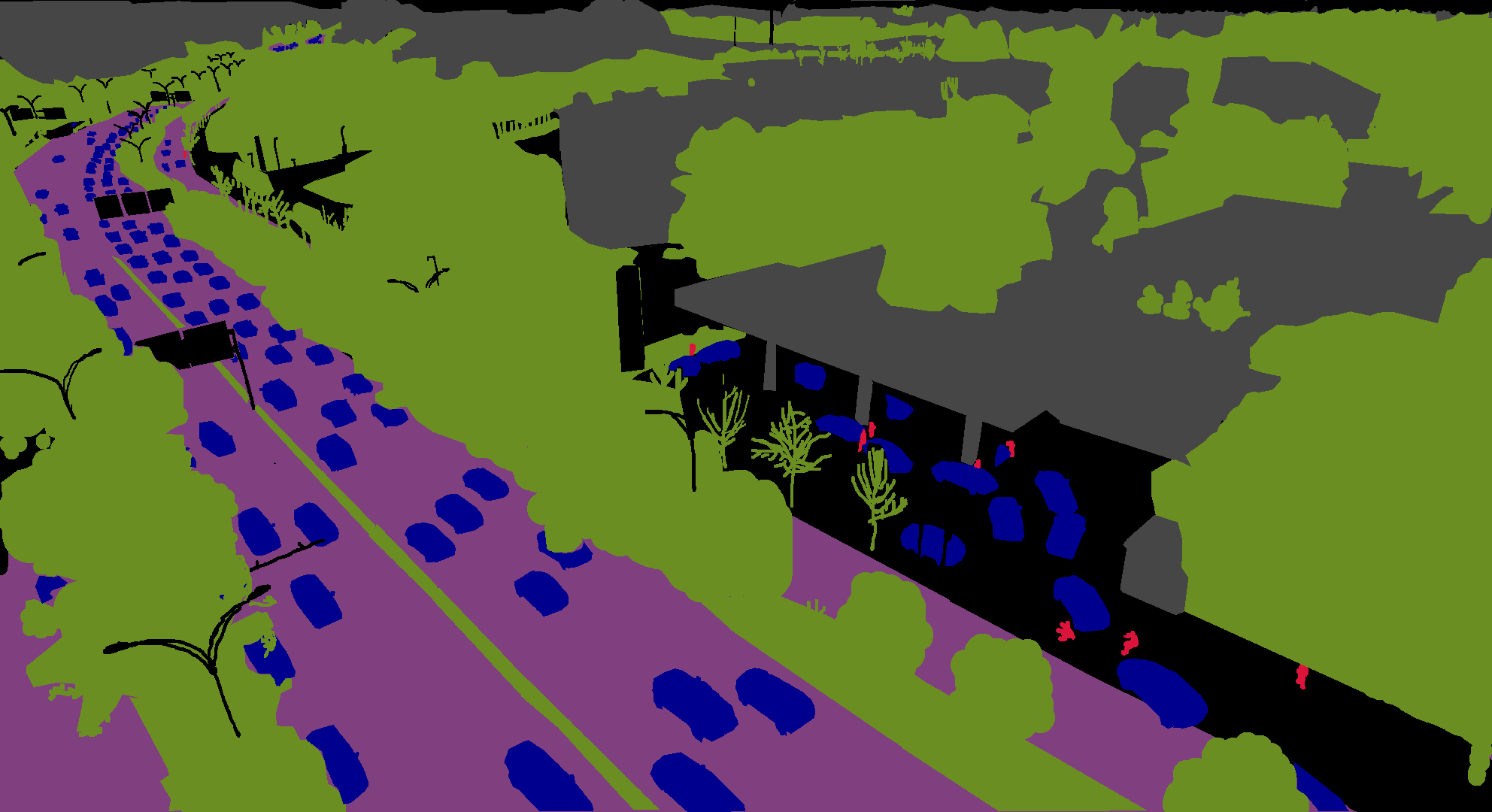}
        \end{subfigure}
        \begin{subfigure}{\textwidth}
            \includegraphics[width=\textwidth,trim=0 15 0 15,clip]{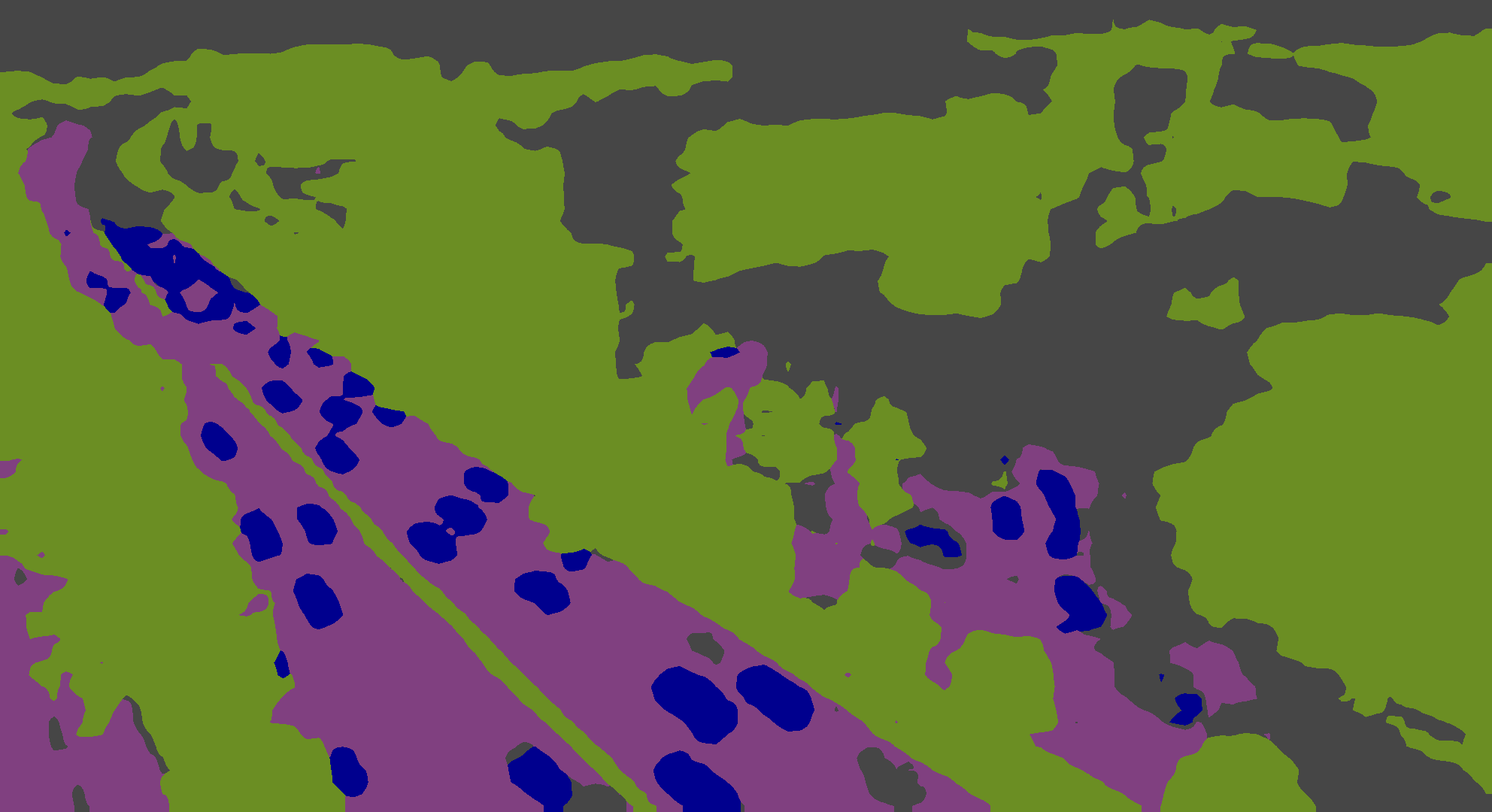}
        \end{subfigure}
    \end{subfigure}
    \begin{subfigure}{\imsize}
        \begin{subfigure}{\textwidth}
            \includegraphics[width=\textwidth]{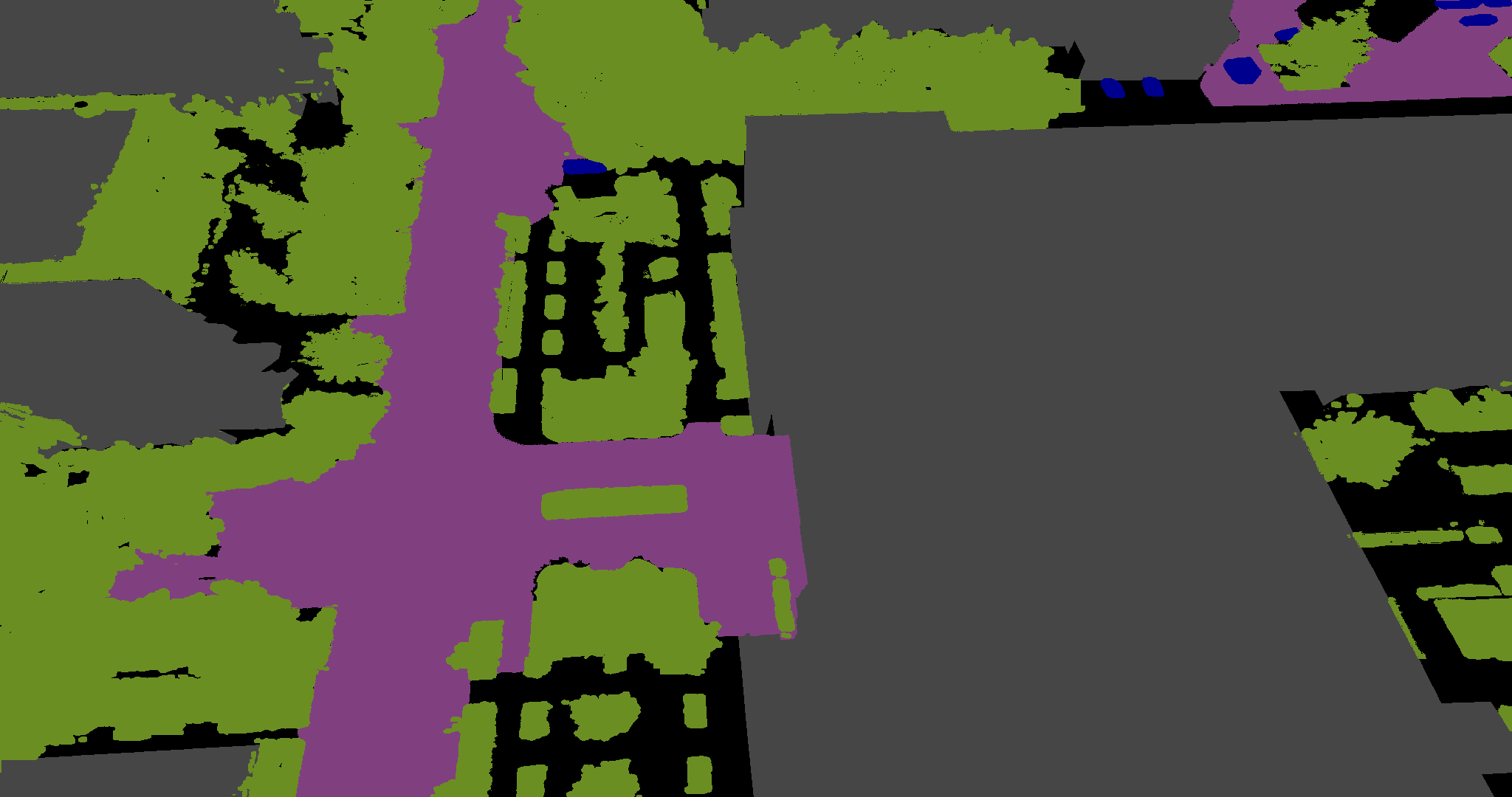}
        \end{subfigure}
        \begin{subfigure}{\textwidth}
            \includegraphics[width=\textwidth]{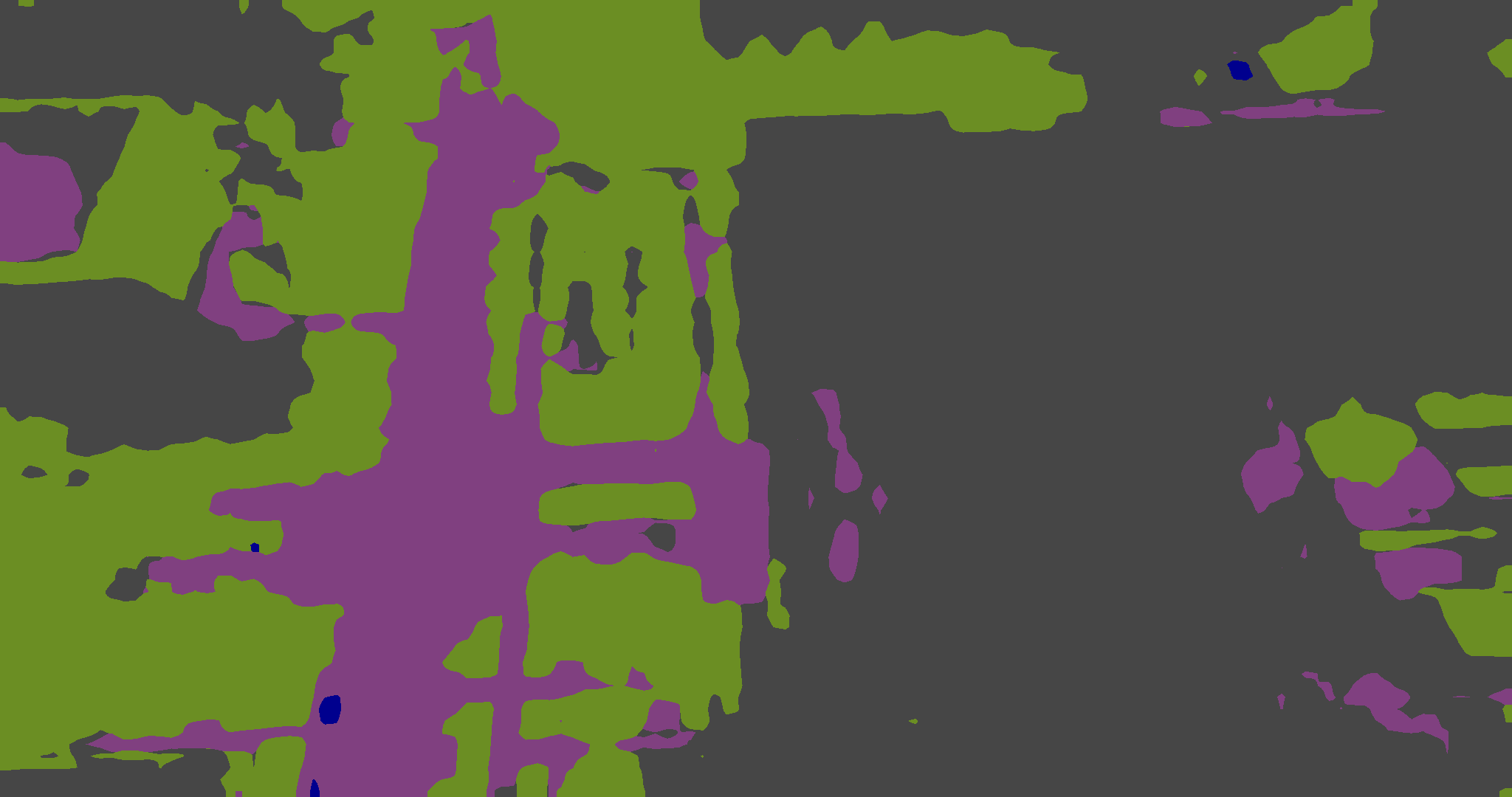}
        \end{subfigure}
    \end{subfigure}
    \begin{subfigure}{\imsize}
        \begin{subfigure}{\textwidth}
            \includegraphics[width=\textwidth]{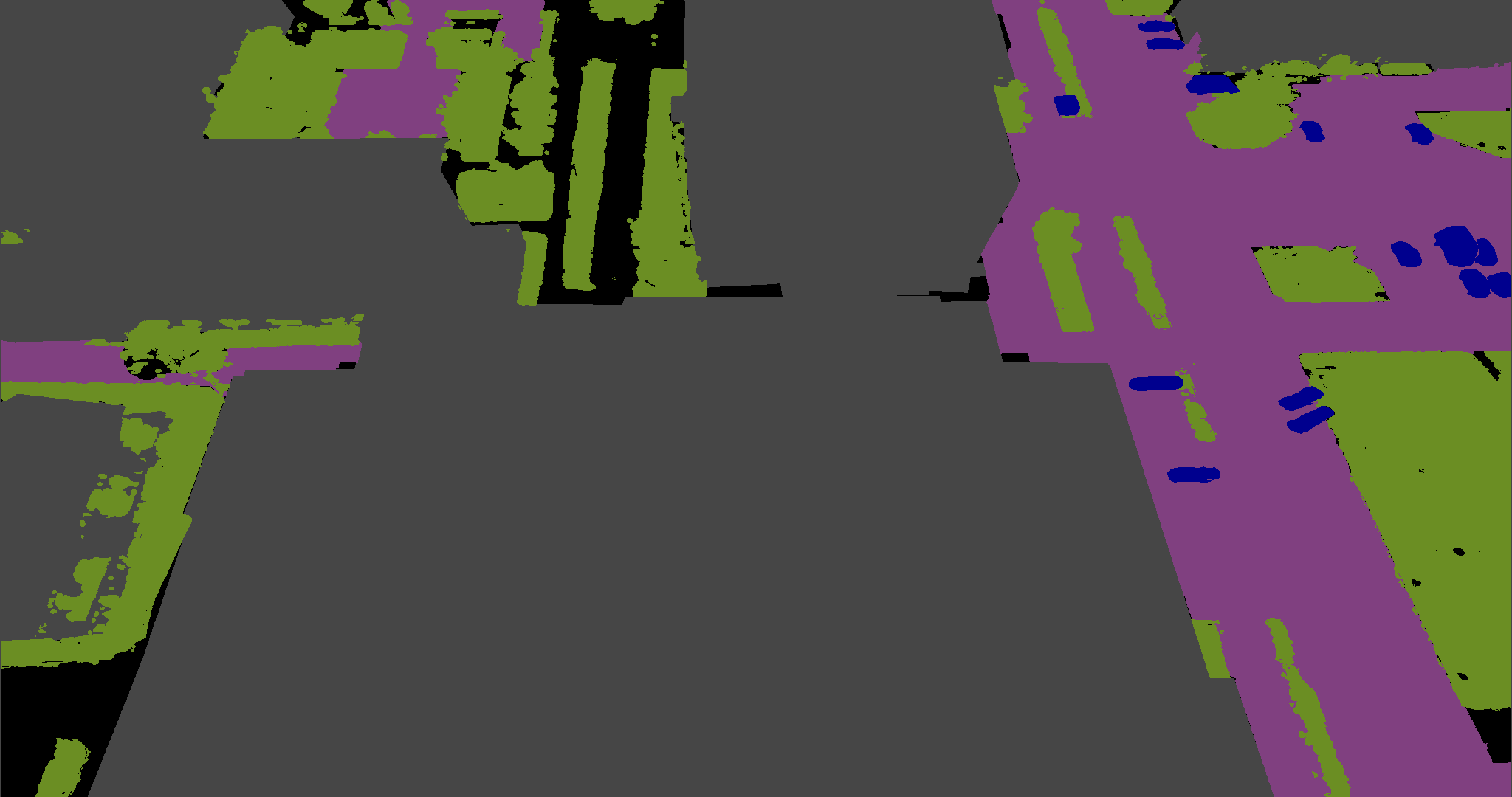}
        \end{subfigure}
        \begin{subfigure}{\textwidth}
            \includegraphics[width=\textwidth]{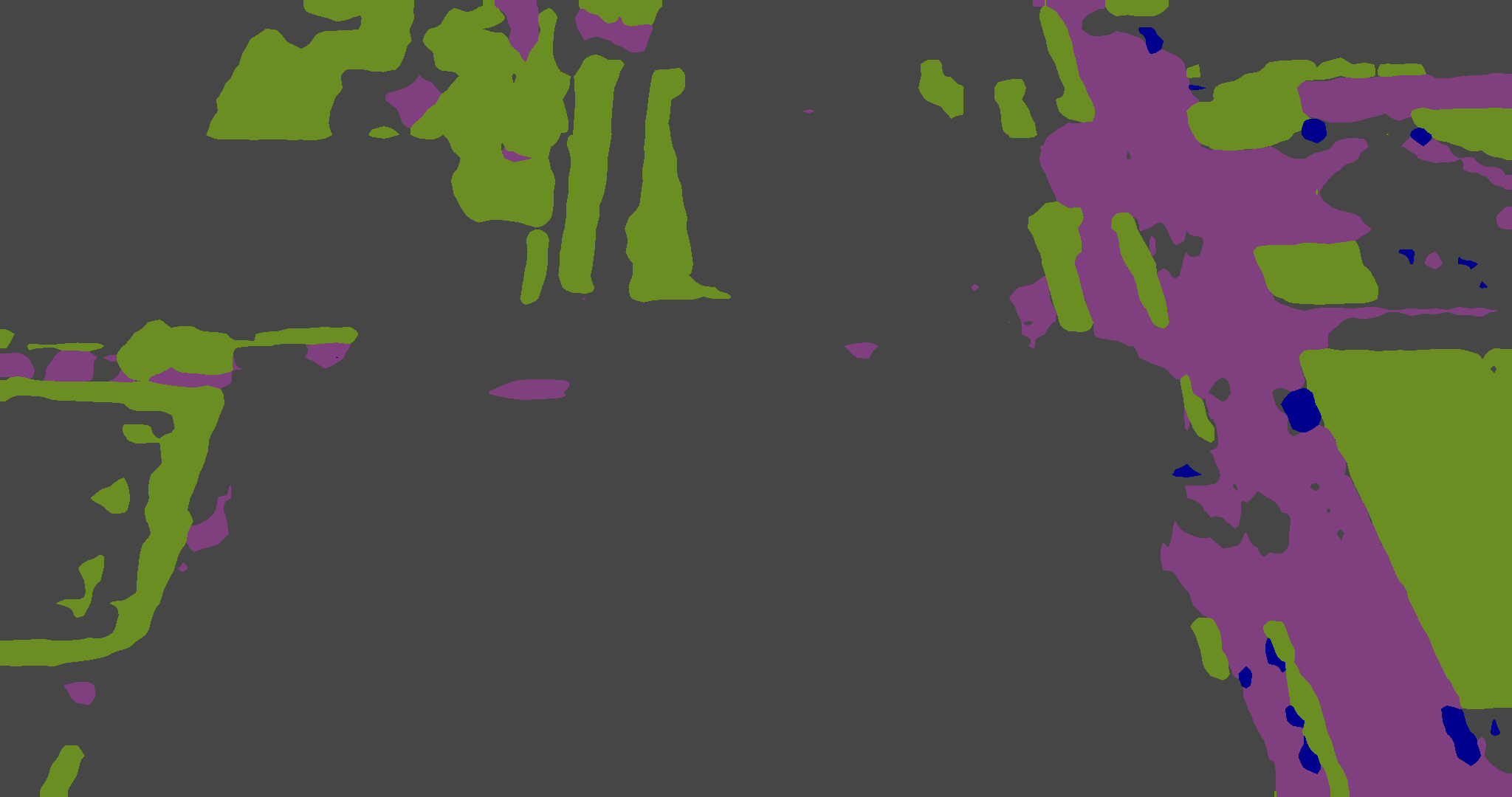}
        \end{subfigure}
    \end{subfigure}
    \begin{subfigure}{\textwidth}
        \begin{tabularx}{\textwidth}{YYYYYYYY}
            \cellcolor{road}\color{white}Road & \cellcolor{nature}Nature & \cellcolor{person}\color{white}Person & \cellcolor{vehicle}\color{white}Vehicle & \cellcolor{construction}\color{white}Construction & \cellcolor{obstacle}Obstacle & \cellcolor{water}\color{white}Water & \cellcolor{unknown}\color{white}{Void}
        \end{tabularx}
    \end{subfigure}
    \caption{Qualitative results of models trained on \dname and tested on real-world data: (GT) Ground-truth semantic map, (Pred) prediction of the model trained on \dname[]. Note that, as generally in semantic segmentation, \textit{void} is ignored during training.}
    \label{fig:qualitative}
\end{figure*}

\subsection*{Evaluation on different towns} 
To further investigate the model's performance, we conducted comparative training and testing across different towns. %
By analyzing Table \ref{tab:towns}, it is evident that the model's accuracy varies significantly across different town scenarios. The highest accuracy is achieved when training and testing are conducted on all towns together, indicating the importance of incorporating more diversified data from a  range of towns for improved generalization. On the other hand, the model's accuracy drops considerably when tested on specific towns that were not part of its training set (notice that not all towns have data for all classes and this can impact severely performances if the training town does not include some of the classes in the test one).
Interestingly, the results also reveal that certain pairs of towns exhibit reasonable performances (\eg, when trained on town ``t01'' and tested on town ``t02'', the model achieves an accuracy of $27.9\%$), while in other cases performances are very low. %
This indicates that there might be similarities or shared visual patterns between some towns, allowing the model to generalize well in these particular cases.
In general, the heterogeneity in performance between towns highlights the need of gathering and combining data from numerous locations in order to increase the model's capacity to generalize well to newly encountered town settings. The model's performance is influenced by the distinct characteristics and scene variations in each town, highlighting the need for comprehensive training datasets that cover diverse town scenarios.

\subsection*{Evaluation on multi-modal segmentation}

Furthermore, we investigate the effectiveness of multi-modal data fusion for the semantic segmentation task. We tested both early and late fusion approaches combining RGB and Depth (D) data (see Table \ref{tab:mm}). 
First of all, by looking at the performances of the two modalities alone, it is possible to notice that the mIoU score for the depth is slightly lower than that of color data. Still, similar results suggest that depth information carries useful semantic information as well.
The early fusion approach combines the RGB and depth data at the input layer (\ie, a 4-channel RGBD input), while the late fusion approach performs the fusion at the output stage, requiring twice the computational cost. In particular, for the late fusion, we replicated the whole architecture (decoder included) and opted to merge the predicted logits using a $1\times1$ convolution, effectively mapping the $2\times C$ channels into $C$ for the final segmentation prediction.
The mIoU score for the latter configuration improves by $3.1\%$ over the RGB results, showing that the multimodal data has a greater information content, although it is important to recognize that it doubles the required operations. Notice that these are just baseline results with naive fusion strategies to offer a starting point for future research. There is a large amount of work on multi-modal segmentation and state-of-the-art strategies will very likely achieve better performances.

\begin{table}[tbph]
    \setlength\tabcolsep{1px}
    \def\arraystretch{1.1}
    \centering
    \resizebox*{1\columnwidth}{!}{
    \begin{tabular}{cccccc}
    \hline
    \textbf{\begin{tabular}{c}Ours\\ (coarse)\end{tabular}} & \textbf{\begin{tabular}{c}Ours\\ (fine)\end{tabular}}& \textbf{Aeroscapes} & \textbf{\begin{tabular}{c}ICG\\ Drone\end{tabular}} & \textbf{UAVid} & \textbf{UDD} \\ \hline
    Road & Road & Road & Paved Area & Road & Road \\
     & Ground &  &  &  &  \\
     & Sidewalk &  &  &  &  \\
     & Road Line &  &  &  &  \\
     & Rail Track &  &  &  &  \\ \hline
    Nature & Vegetation & Vegetation & Vegetation & Vegetation & Vegetation \\
     & Terrain &  & Tree & Tree &  \\
     &  &  & Grass &  &  \\
     &  &  & Dirt &  &  \\
     &  &  & Gravel &  &  \\
     &  &  & Rocks &  &  \\ \hline
    Person & Person & Person & Person & Human &  \\ \hline
    Vehicle & Car & Car & Car & Static Car & Vehicle \\
     & Truck & Bicycle & Bicycle & Dynamic Car &  \\
     & Bus &  &  &  &  \\
     & Train &  &  &  &  \\
     & Motorcycle &  &  &  &  \\
     & Bicycle &  &  &  &  \\ \hline
     Construction & Building & Construction & Roof & Building & Roof \\ 
     & Wall &  & Wall &  & Facade \\
     & Fence &  & Fence &  &  \\
     & Bridge &  & Window &  &  \\
     &  &  & Door &  &  \\
     &  &  & Fence Pole &  &  \\ \hline
     Obstacle & Other & Obstacle & Obstacle &  &  \\
     & Pole &  &  &  &  \\
     & Traffic Signs &  &  &  &  \\
     & Guard Rail &  &  &  &  \\
     & Traffic Light &  &  &  &  \\
     & Static &  &  &  &  \\
     & Dynamic &  &  &  &  \\ \hline
    Water & Water &  & Water &  &  \\
     &  &  & Pool &  & \\
     \hline
     \end{tabular}
    }
    \caption{Coarse class re-mapping for synthetic-to-real adaptation.}
    \label{tab:mapping}
\end{table}
\subsection{Synthetic-to-real training}
A key aspect in the evaluation of the quality of a synthetic dataset is the capability of models learned on it to perform well on real-world data.
Aiming to perform this evaluation on state-of-the-art datasets, we performed a re-mapping of the labels into a common 8 classes set, Table \ref{tab:mapping} shows how the labels in the different datasets are mapped to our common set.
In Table \ref{tab:adapt} we show the performances of the same model, \ie, DeeplabV3 with MobilenetV3, trained and tested on different datasets.
The \textit{Oracle} tests, which assume training and testing on the same dataset, use the same set of parameters as the previous tests with some modifications. Due to the limited size of the datasets, the tests were conducted with 30k iterations to prevent overfitting. Additionally, for all datasets with resolutions ranging from 2-4k, we downscaled the data to full HD while maintaining the original aspect ratio. This adjustment was necessary to ensure compatibility with the network architecture. Notably, for the ICG Drone dataset, since no training and testing splits were provided, the test has not been performed (as such the metric reported for the oracle is basically the training accuracy, which is an overestimation of the performance). %
Generally, our dataset, without any augmentation or adaptation (\ie, performing \textit{source-only} training), demonstrates good generalization performance across the majority of datasets. However, it faces challenges when tested on more complex datasets, where the accuracy of class mapping is less precise. It is worth mentioning a particular case, \ie, ICG Drone, where the absence of the road class and a focus on non-urban areas, mainly green and residential zones, affect the results. Nevertheless, the model trained on our dataset still achieves promising results in these scenarios, and there is potential for further enhancement by exploring transfer learning and domain adaptation techniques.
In Figure \ref{fig:qualitative}, the qualitative results of the trained model on the real-world data are shown. %
The reconstruction of semantic maps remains unaffected by factors such as height, viewing angle, or variations in traffic density, encompassing both heavy traffic and sparsely populated roads.

\section{Conclusion}

In this paper, we introduced a new multimodal synthetic dataset for UAVs, focusing on the costly and scarcely available densely-annotated data. The dataset contains several sequences recorded in different synthetic towns and with a multimodal sensor array, providing ground truth depths, semantic maps, 3D bounding boxes, and semantic LiDAR information.
Given the heterogeneous nature of recording heights found in real datasets, we opted to render our samples from three different altitudes (20m, 50m, and 80m) with different camera orientations. %
In total, our dataset offers 72k samples with pixel-level annotations split into 60k training samples and 12k test samples. 
We provide multiple benchmark results for semantic segmentation and object detection by training standard networks on our dataset.
Additionally, we performed some  studies on the generalization capability of the trained architectures when tested on the presence of domain shift (town$\rightarrow$town and height$\rightarrow$height), highlighting the need for heterogeneous data during training.
We also investigate the generalization potential of our dataset in the synthetic-to-real scenario, testing a model trained on our dataset on different real datasets without any explicit adaptation strategies, achieving results that clearly show the potential of the dataset in the task.

In the future, we plan to further extend the dataset including more sensors and a bigger variety of settings. Domain adaptation strategies will be also tested in order to better evaluate the generalization capabilities of the dataset.

{\small
\bibliographystyle{IEEEtran}
\bibliography{egbib}
}

\end{document}